\documentclass[final,5p,times,twocolumn]{elsarticle}

\usepackage{amssymb}
\usepackage{amsmath}
\usepackage{subcaption}
\usepackage{booktabs}
\usepackage{multirow}
\usepackage{lineno}

\journal{Computers and Chemical Engineering}

\begin{document}

\begin{frontmatter}



\title{Comparative Evaluation of Kolmogorov–Arnold Autoencoders and Orthogonal Autoencoders for Fault Detection with Varying Training Set Sizes}


\author[label1]{Enrique Luna Villagómez}

\author[label1]{Vladimir Mahalec\corref{cor1}}
\ead{mahalec@mcmaster.ca}
\cortext[cor1]{Corresponding author}

\affiliation[label1]{organization={McMaster University, Department of Chemical Engineering}, 
            addressline={1280 Main St W}, 
            city={Hamilton},
            postcode={L8S 4L8}, 
            state={Ontario},
            country={Canada}}

\begin{abstract}
Kolmogorov–Arnold Networks (KANs) have recently emerged as a flexible and parameter-efficient alternative to conventional neural networks. Unlike standard architectures that use fixed node-based activations, KANs place learnable functions on edges, parameterized by different function families. While they have shown promise in supervised settings, their utility in unsupervised fault detection remains largely unexplored. This study presents a comparative evaluation of KAN-based autoencoders (KAN-AEs) for unsupervised fault detection in chemical processes. We investigate four KAN-AE variants, each based on a different KAN implementation (EfficientKAN, FastKAN, FourierKAN, and WavKAN), and benchmark them against an Orthogonal Autoencoder (OAE) on the Tennessee Eastman Process. Models are trained on normal operating data across 13 training set sizes and evaluated on 21 fault types, using Fault Detection Rate (FDR) as the performance metric. WavKAN-AE achieves the highest overall FDR ($\geq$92\%) using just 4,000 training samples and remains the top performer, even as other variants are trained on larger datasets. EfficientKAN-AE reaches $\geq$90\% FDR with only 500 samples, demonstrating robustness in low-data settings. FastKAN-AE becomes competitive at larger scales ($\geq$50,000 samples), while FourierKAN-AE consistently underperforms. The OAE baseline improves gradually but requires substantially more data to match top KAN-AE performance. These results highlight the ability of KAN-AEs to combine data efficiency with strong fault detection performance. Their use of structured basis functions suggests potential for improved model transparency, making them promising candidates for deployment in data-constrained industrial settings.

\end{abstract}




\end{frontmatter}



\section{Introduction}
\label{sec:intro}
Fault detection and diagnosis (FDD) remains a critical task in modern industrial systems. Faults such as equipment malfunctions, sensor degradation, and control system anomalies can compromise operational safety, reduce productivity, and lead to environmental harm. Although faults cannot always be prevented, process monitoring systems can detect deviations from normal operation and enable timely intervention~\cite{WANG2009,KOPBAYEV2022}. Designing FDD methods that perform reliably under multivariate, dynamic, and drifting conditions constitutes a central challenge in industrial process monitoring~\cite{Park2020}. The rise of Industry 4.0 has led to the widespread deployment of sensors across process units, significantly increasing the availability of high-resolution multivariate data~\cite{KAUPP2021}. As a result, there has been a shift from model-based techniques to data-driven frameworks, which use statistical or machine learning methods to infer system behavior directly from process measurements~\cite{Sun2020}. Data-driven frameworks span a wide range of approaches, from classical methods such as Principal Component Analysis (PCA)~\cite{Tamura2007,Ge2013,Luna2025} to more flexible models based on artificial neural networks (ANNs), including multilayer perceptrons (MLPs)~\cite{Cacciarelli2022,Zhu2021}, convolutional neural networks (CNNs)~\cite{Ren2019,Lomov2021}, and recurrent neural networks (RNNs)~\cite{KANG2020,Sun2020}.

Despite their empirical success, ANN-based methods are often criticized for their lack of interpretability, as the internal mechanisms governing their predictions are difficult to analyze. Even with post hoc techniques such as saliency mapping~\cite{simonyan2014}, Layer-wise Relevance Propagation (LRP)~\cite{bach2015}, or SHAP~\cite{Lundberg2017}, understanding the contribution of individual features remains challenging. Furthermore, these models often require access to large volumes of labeled or representative data to achieve reliable performance, which may not be feasible in practice. In response to these limitations, Liu et al.~\cite{liu2025} introduced Kolmogorov–Arnold Networks (KANs) as a more interpretable and parameter-efficient alternative to traditional MLPs. In a KAN, each connection is modeled by a learnable univariate function, and each node computes its output as the sum of its transformed inputs~\cite{ji2025}. The explicit definition of each univariate function promotes both interpretability and parameter efficiency. These characteristics make KANs especially suitable for fault detection tasks that demand transparent decision-making under limited data availability.

KANs have recently been applied to a range of supervised FDD tasks. For instance, Rigas et al.~\cite{Rigas2025} developed shallow KAN-based classifiers for diagnosing rotating machinery faults, achieving high F1 scores while using significantly fewer parameters than standard deep networks. Similarly, Cabral et al.~\cite{Cabral2025} introduced KANDiag for diagnosing faults in oil-immersed power transformers under imbalanced conditions, showing improved robustness over established diagnostic standards and machine learning baselines. Other works have proposed hybrid frameworks that integrate deep feature extraction with KAN-based decision layers. He and Mo~\cite{He2025} presented WCNN-KAN, which extracts wavelet-based time-frequency representations of vibration signals before classification using a KAN. Zhang et al.~\cite{ZHANG2025} proposed Conv-KAN for refrigerant fault diagnosis in variable refrigerant flow (VRF) air conditioning systems, demonstrating improved accuracy and faster convergence relative to MLP and CNN baselines.

Although recent applications of KANs to fault detection have shown promising results, existing work is limited in scope. First, most studies rely on the original KAN formulation, which uses B-spline basis functions to parameterize the learnable activation functions. The effect of alternative basis function choices on detection performance has not been systematically assessed. Since basis functions impose distinct inductive biases (e.g., smoothness, locality, frequency resolution), they may significantly impact detection performance across different fault types. Second, nearly all current studies assume access to labeled fault data. This excludes unsupervised scenarios where only normal operating data are available, despite the practical relevance of such scenarios in industrial settings. Third, it remains unclear whether the parameter efficiency of KANs is preserved when training data are scarce, or how sensitive their performance is to the amount of training data. The robustness of KANs under varying training data availability has yet to be fully characterized.

To address these limitations, this study presents a comprehensive evaluation of Kolmogorov-Arnold Autoencoders (KAN-AE) for unsupervised fault detection in chemical process systems. We investigate four KAN variants (EfficientKAN, FastKAN, FourierKAN, and WavKAN), each employing a distinct basis function family (e.g., B-splines, wavelets, Fourier series) to construct the univariate edge functions that define the network’s nonlinear transformations. Our contributions are as follows: (1) we assess the data efficiency of KAN-AEs by analyzing how fault detection performance scales with training set size; (2) we evaluate the influence of functional parameterization on fault sensitivity across diverse fault categories; and (3)  we perform statistically grounded model comparisons using Bayesian signed-rank tests, estimating posterior probabilities of model superiority and practical equivalence across fault scenarios. 

The remainder of this paper is organized as follows: Section~\ref{sec:preliminaries} introduces the theoretical background on autoencoders and Kolmogorov–Arnold Networks, including architectural variants and mathematical foundations. Section~\ref{sec:methodology} details the experimental setup, including the Tennessee Eastman Process benchmark, model training procedures, fault detection strategy, and performance evaluation. Section~\ref{sec:results and discussion} presents and analyzes the experimental results. Finally, Section~\ref{sec:conclusions} summarizes the key findings and outlines implications for industrial fault detection and future research directions.

\section{Preliminaries}
\label{sec:preliminaries}
In this section, the modeling approaches considered in the present comparative study, including autoencoders and Kolmogorov-Arnold Networks, are briefly reviewed.

\subsection{Autoencoders}
Autoencoders (AEs) are neural networks that learn low-dimensional representations of high-dimensional data by reconstructing inputs from compressed latent variables. Unlike linear techniques such as Principal Component Analysis (PCA), Independent Component Analysis (ICA), or Partial Least Squares (PLS), AEs can model nonlinear transformations, making them particularly suitable for capturing the complex variable interactions typical of chemical processes \cite{KONG2022}. An AE consists of an encoder–decoder pair trained to minimize a reconstruction loss,  defined as the mean squared error between the input samples and their reconstructions:

\begin{equation}
\label{eq:loss_ae}
L(x, \hat{x}) = \frac{1}{N} \sum_{i=1}^{N} \left\| x_i - g(f(x_i)) \right\|_2^2.
\end{equation}
\noindent
where \( f(\cdot) \) and \( g(\cdot) \) denote the encoder and decoder functions, respectively. Despite their representational flexibility, standard AEs often learn correlated latent features, which degrades fault detection performance due to poor covariance modeling. To address this, the Orthogonal Autoencoder (OAE) \cite{Cacciarelli2022} introduces a regularization term that encourages decorrelation among latent dimensions:

\begin{equation}
\label{eq:loss_oae}
\mathcal{L}_{\text{OAE}}(x, \hat{x}) = \frac{1}{N} \sum_{i=1}^{N} \left\| x_i - g(f(x_i)) \right\|_2^2 + \lambda \left\| Z^\top Z - I \right\|_F^2,
\end{equation}
\noindent
where \( Z \in \mathbb{R}^{N \times k} \) is the matrix of latent representations for a mini-batch of \( N \) samples and \( k \) latent features, \( I \in \mathbb{R}^{k \times k} \) is the identity matrix, and \( \lambda \) controls the strength of the  regularization. The OAE serves as a baseline in this study for evaluating the advantages and limitations of KAN-AE architectures.

\subsection{Kolmogorov-Arnold Networks}
Kolmogorov–Arnold Networks (KANs) are a recent neural architecture designed to offer a more expressive and interpretable alternative to traditional multilayer perceptrons  \cite{liu2025}. They are inspired by the Kolmogorov–Arnold representation theorem, which states that any continuous multivariate function defined on a bounded domain can be expressed as a finite composition of continuous univariate functions and additions \cite{Kolmogorov1957}. This theoretical result motivates the core design of KANs: rather than using fixed scalar weights as in conventional neural networks, each edge in a KAN is parameterized by a learnable univariate function, enabling the model to approximate complex nonlinear relationships with enhanced adaptability and transparency.

In a KAN, each layer transforms its input by applying a set of learnable univariate functions to individual input features and then summing their outputs. Specifically, each edge from neuron \( j \) in layer \( \ell - 1 \) to neuron \( i \) in layer \( \ell \) is associated with a function \( \phi_{ij}^{(\ell)} \colon \mathbb{R} \rightarrow \mathbb{R} \). The output of neuron \( i \) is computed as:

\begin{equation}
x_i^{(\ell)} = \sum_{j=1}^{n_{\ell-1}} \phi_{ij}^{(\ell)}(x_j^{(\ell-1)}).
\end{equation}
\noindent
where \( \phi_{ij}^{(\ell)} \colon \mathbb{R} \rightarrow \mathbb{R} \) is parameterized using a smooth basis expansion. This edge-wise functional form offers two benefits critical for process monitoring: high representational capacity and greater transparency in how input variables influence outputs. Several architectural variants of KANs have been proposed, differing primarily in the basis family used to parameterize the univariate functions. We review four representative variants below.

\subsection{KAN Variants}
The functions \( \phi_{ij}^{(\ell)} \) in a KAN can be parameterized using different basis families, such as B-splines, Gaussian RBFs, Fourier modes, and wavelets. In this study, we evaluate how these basis choices influence the performance of KAN-AE variants in fault detection tasks.

\subsubsection{EfficientKAN}
EfficientKAN is a memory-efficient KAN variant designed to address the scalability bottlenecks of the original formulation \cite{blealtan2024}. In the original implementation \cite{liu2025}, each function \( \phi_{ij}^{(\ell)} \) is parameterized as a weighted combination of a fixed base activation and a B-spline expansion:

\begin{equation}
\label{eq:KAN-Bspline}
\phi_{ij}^{(\ell)}(x)= w_{ij}^{(\ell)}\,b(x)+
s_{ij}^{(\ell)}\sum_{k=1}^{g}\theta_{ijk}^{(\ell)}\,B_k^{(\ell)}(x),
\end{equation}
\noindent
where $b(x)$ is included to stabilize training and improve convergence, \( B_k^{(\ell)}(x) \) are B-spline basis functions of order \( r \), and \( w_{ij}^{(\ell)} \), \( s_{ij}^{(\ell)} \), and \( \theta_{ijk}^{(\ell)} \) are learnable parameters. Although this per-edge B-spline formulation enables the network to flexibly capture localized nonlinearities, it requires evaluating a distinct transformation for every input–output edge. This results in large intermediate tensors and high memory usage, particularly in deep networks or large mini-batches. EfficientKAN mitigates this issue by applying the B-spline basis once per input feature and reusing the resulting activations across all outputs via matrix multiplication, significantly improving memory and computational efficiency.

In EfficientKAN, the L1 and entropy regularization terms proposed by Liu et al. \cite{liu2025} are approximated using the average absolute value of the spline coefficients, which serves as a proxy for the activity of each edge. Specifically, for edge \( (i,j) \) in layer \( \ell \), activity is estimated as:

\begin{equation}
\bar{\theta}_{ij}^{(\ell)} = \frac{1}{g} \sum_{k=1}^{g} \left| \theta_{ijk}^{(\ell)} \right|,
\end{equation}

\noindent
and the total activity in the layer is given by:

\begin{equation}
\label{eq:l1_activity}
\bar{\Theta}^{(\ell)} = \sum_{i=1}^{n_\ell} \sum_{j=1}^{n_{\ell-1}} \bar{\theta}_{ij}^{(\ell)}.
\end{equation}
\noindent
The entropy penalty is then computed as:

\begin{equation}
\label{eq:entropy_activity}
S(\Phi^{(\ell)}) = - \sum_{i=1}^{n_\ell} \sum_{j=1}^{n_{\ell-1}} 
\frac{ \bar{\theta}_{ij}^{(\ell)} }{ \bar{\Theta}^{(\ell)} }
\log \left( \frac{ \bar{\theta}_{ij}^{(\ell)} }{ \bar{\Theta}^{(\ell)} } \right).
\end{equation}

\subsubsection{FastKAN}
FastKAN is a computationally streamlined KAN variant that improves forward-pass efficiency by replacing the adaptive B-spline basis with a fixed set of Gaussian radial basis functions (RBFs) \cite{Li2024}. Each function \( \phi_{ij}^{(\ell)} \colon \mathbb{R} \to \mathbb{R} \) is expressed as a combination of a base activation $b(x)$ and a linear combination of \( g \) Gaussian RBFs centered at fixed locations \( \mu_k \), with common width \( \sigma \):

\begin{equation}
\phi_{ij}^{(\ell)}(x) = w_{ij}^{(\ell)} \cdot b(x) + \sum_{k=1}^{g} \theta_{ijk}^{(\ell)} \exp\left( -\left( \frac{x - \mu_k}{\sigma} \right)^2 \right),
\end{equation}
\noindent
where \( w_{ij}^{(\ell)} \) and \( \theta_{ijk}^{(\ell)} \) are trainable parameters. Unlike B-spline-based KANs, the basis is fixed and not adapted during training. This reduces computational complexity and supports scalability, while retaining per-edge functional flexibility. To keep inputs within the RBF support region, a layer normalization step is often applied before the RBF transformation.

\subsubsection{FourierKAN}
FourierKAN replaces localized basis expansions with globally supported, periodic functions by parameterizing each learnable function \( \phi_{ij}^{(\ell)} \colon \mathbb{R} \to \mathbb{R} \) using a truncated Fourier series \cite{xu2024}:

\begin{equation}
\phi_{ij}^{(\ell)}(x) = \sum_{k=1}^{g} \left( \alpha_{ijk}^{(\ell)} \cos(k x) + \beta_{ijk}^{(\ell)} \sin(k x) \right) + b_{ij}^{(\ell)},
\end{equation}
\noindent
where \( g \) denotes the number of Fourier modes, \( \alpha_{ijk}^{(\ell)} \), \( \beta_{ijk}^{(\ell)} \), and \( b_{ij}^{(\ell)} \) are learnable parameters. The globally supported Fourier basis simplifies implementation by eliminating knot management and promotes stable training through smoother gradients.

\subsubsection{WavKAN}
WavKAN is a KAN variant proposed by Bozorgasl and Chen \cite{bozorgasl2024}, in which each learnable function \( \phi_{ij}^{(\ell)} \colon \mathbb{R} \to \mathbb{R} \) is parameterized as a scaled and shifted instance of a single wavelet:

\begin{equation}
\phi_{ij}^{(\ell)}(x) = w_{ij}^{(\ell)} \psi\left( \frac{x - t_{ij}^{(\ell)}}{s_{ij}^{(\ell)}} \right),
\end{equation}
\noindent
where \( w_{ij}^{(\ell)} \), \( t_{ij}^{(\ell)} \), and \( s_{ij}^{(\ell)} \) are trainable parameters controlling amplitude, translation, and scale, respectively. This direct parameterization eliminates the need for explicit basis expansions or adaptive knot placement, as required in B-spline-based KANs. By assigning one wavelet per edge, WavKAN supports localized adaptation to both low- and high-frequency patterns. This design enables multiresolution representations, allowing the model to capture sharp variations and broader trends simultaneously, which is advantageous for modeling structured signals in industrial process data.

\begin{figure}[t]
    \centering
    \begin{subfigure}[t]{0.48\textwidth}
        \centering
        \includegraphics[width=0.8\textwidth]{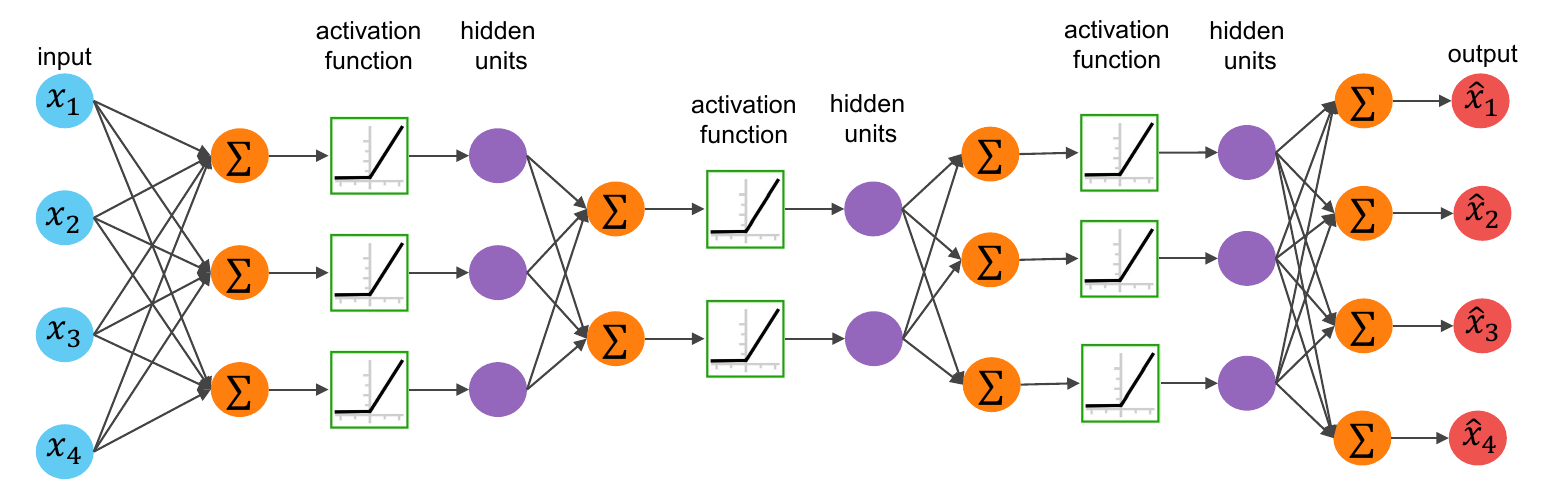}
        \caption{MLP-based autoencoder.}
        \label{fig:MLP-AE-a}
    \end{subfigure}
    
    \vspace{1em} 
    
    \begin{subfigure}[t]{0.48\textwidth}
        \centering
        \includegraphics[width=0.8\textwidth]{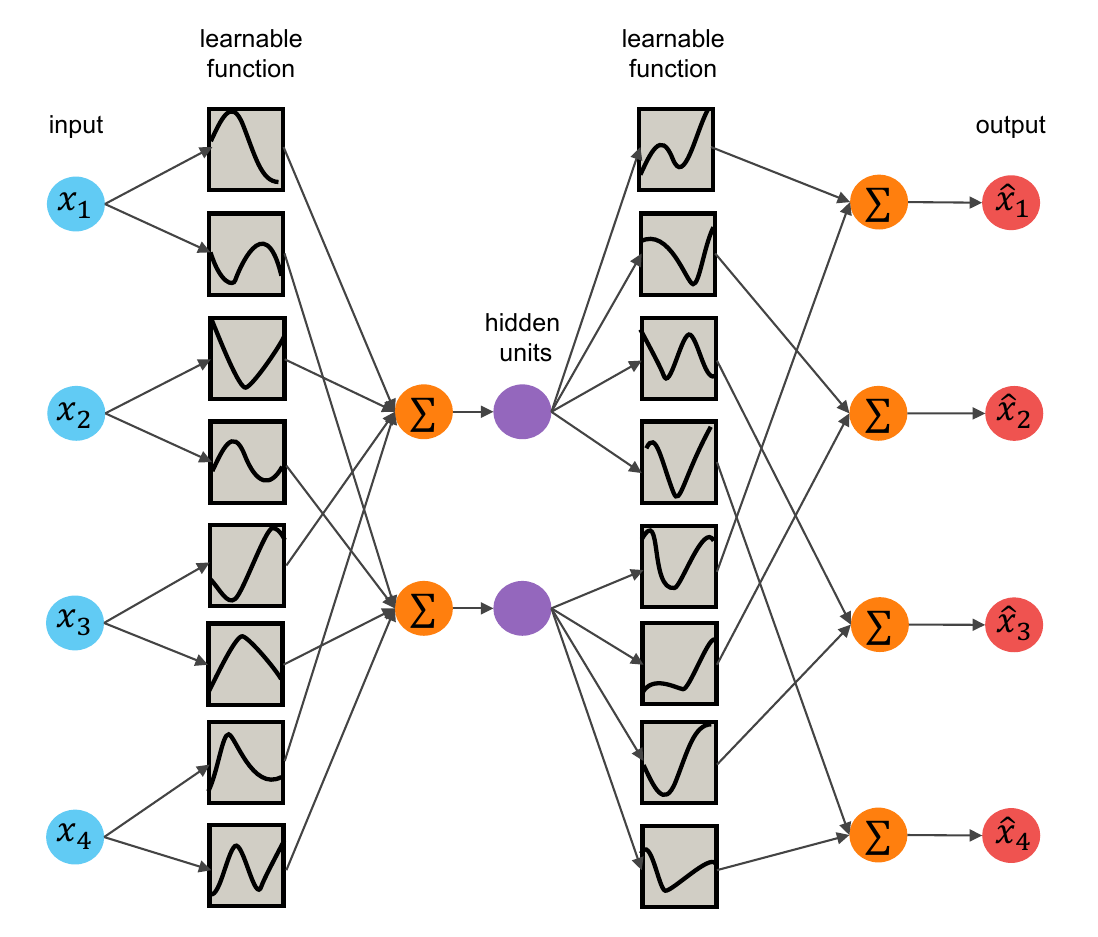}
        \caption{KAN-based autoencoder.}
        \label{fig:KAN-AE-b}
    \end{subfigure}
    
    \caption{Comparison of (a) MLP-based and (b) KAN-based autoencoders. KAN-based models replace fixed activation functions with learnable univariate transformations along each edge.}
    \label{fig:AE-comparison}
\end{figure}

\subsection{Kolmogorov-Arnold Autoencoder}
KAN-based autoencoders (KAN-AEs) extend standard autoencoder architectures (see Fig. \ref{fig:MLP-AE-a}) by replacing the MLP components in the encoder-decoder pair with KAN layers. Unlike MLPs, which apply linear transformations followed by fixed activations at each node, KANs assign a learnable univariate function to each edge. This edge-wise functional parameterization allows KAN-AEs to model sharp local nonlinearities without resorting to deep or wide MLP stacks. Furthermore, it facilitates the interpretation of variable interactions, which may support fault diagnosis in process monitoring tasks.

KAN-AEs can be constructed entirely from KAN layers \cite{yu2024}, or using hybrid architectures that combine KAN and MLP layers \cite{Moradi2024}. In this work, all KAN-AEs are implemented using only KAN layers (see Fig. \ref{fig:KAN-AE-b}). This configuration isolates the effect of functional parameterization, enabling direct attribution of any performance or interpretability gains to the KAN architecture.

\begin{figure}[t]
    \centering
    \includegraphics[angle=90,width=0.48\textwidth]{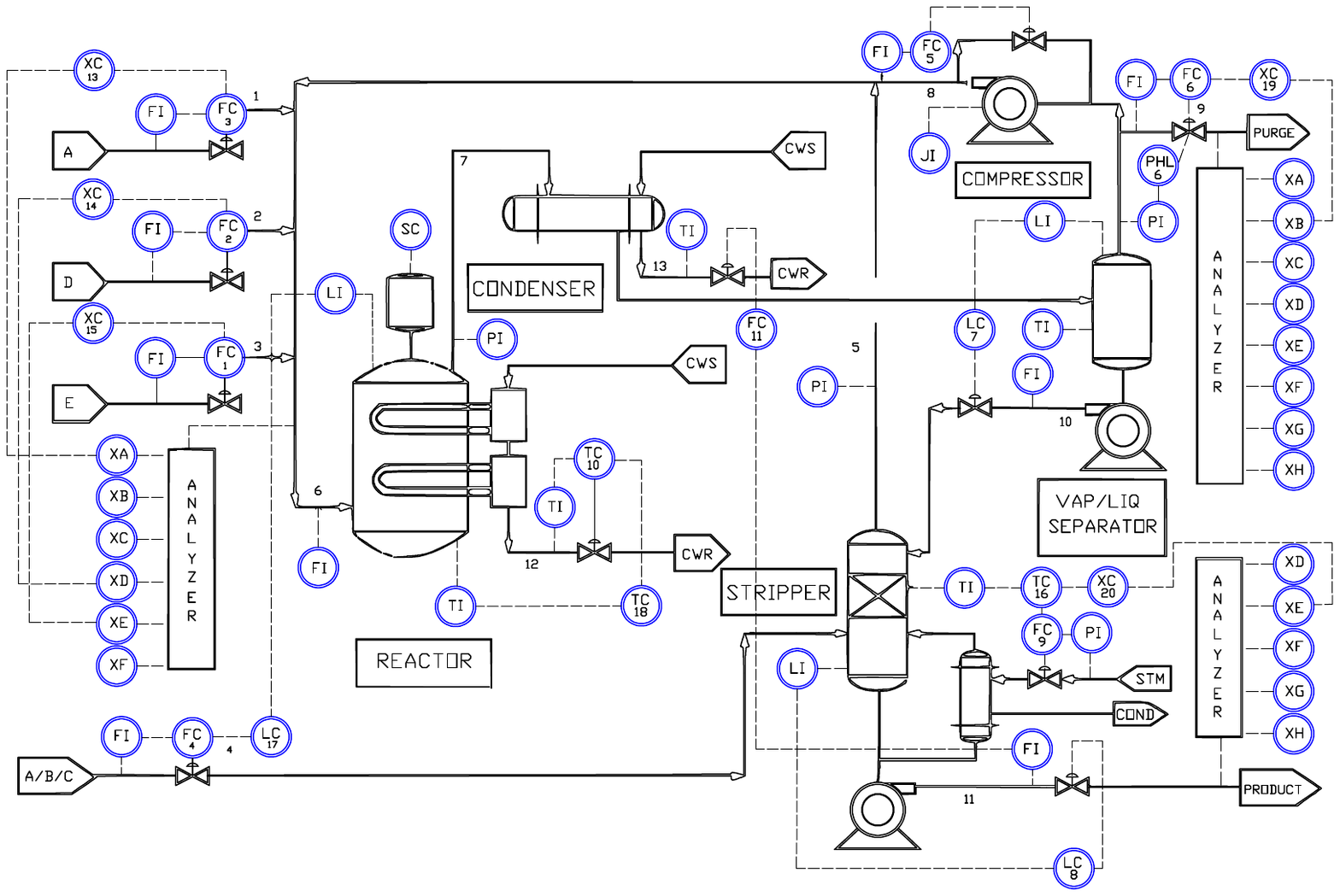}
    \caption{Tennessee Eastman Process flowsheet.}
    \label{fig:TEP_diagram}
\end{figure}

\section{Methodology}
\label{sec:methodology}
This section details the end-to-end pipeline used to configure, train, and evaluate the autoencoder models considered in this study. The primary goal is to compare conventional and KAN-based parameterizations for unsupervised fault detection under matched conditions. To this end, we systematically investigate three core aspects: (1) how detection performance scales with training set size; (2) how functional basis choices in KANs affect generalization across fault types; and (3) the robustness of model comparisons, quantified by the posterior probability that one model consistently outperforms another across fault scenarios. All experiments are conducted using the Tennessee Eastman Process benchmark, which, in the absence of real plant data, provides a diverse and widely adopted suite of fault scenarios.

\subsection{Case Study: The Tennessee Eastman Process}
\label{sec:TEP}
The Tennessee Eastman Process (TEP), introduced by Downs and Vogel \cite{Downs1993}, represents a chemical production system involving eight components and five process units as shown in Fig. \ref{fig:TEP_diagram}. The process involves two primary exothermic reactions ($\mathrm{A} + \mathrm{C} + \mathrm{D} \rightarrow \mathrm{G}$ and $\mathrm{A} + \mathrm{C} + \mathrm{E} \rightarrow \mathrm{H}$) and two secondary byproduct reactions ($\mathrm{A} + \mathrm{E} \rightarrow \mathrm{F}$ and $3\mathrm{D} \rightarrow 2\mathrm{F}$), all occurring in the presence of an inert component ($\mathrm{B}$).

Several implementations of the TEP have been developed based on different control strategies \cite{McAvoy1994, Lyman1995, Ricker1996} and modeling approaches \cite{Russell2000, VILLALBA2018}, resulting in multiple publicly available datasets for research. In this work, we adopt the dataset generated by Rieth et al. \cite{Rieth2017}, which includes a large number of simulation runs, enabling a systematic evaluation of KAN-AEs for fault detection across varying amounts of training data.

The training set consists of 500 simulation runs under normal operating conditions, each containing 500 samples. The testing set includes 500 runs for each of the 21 faults introduced by Russell et al.~\cite{Russell2000}, with measurements recorded every three minutes over a 48-hour period (960 samples per run). The first 160 samples correspond to normal operation, and the remaining 800 samples to abnormal operation. Each dataset contains 11 manipulated variables, $\mathrm{XMV}(1)$–$\mathrm{XMV}(11)$, and 41 process measurements, $\mathrm{XMEAS}(1)$–$\mathrm{XMEAS}(41)$. 

Understanding the nature of the faults is critical for evaluating detection performance, as a high FDR does not necessarily imply effective monitoring. Sun et al. \cite{Sun2020} proposed classifying TEP faults into three categories: controllable, back-to-control, and uncontrollable. In controllable faults, all measured and manipulated variables remain within their normal operating limits despite the presence of disturbances. In back-to-control faults, the measured variables are restored to their normal ranges, but at least one manipulated variable deviates from its allowable range. In uncontrollable faults, neither the measured nor the manipulated variables can be maintained within their normal operating boundaries despite the actions of the control system.

The benchmark dataset includes 21 faults covering diverse disturbance types, including step changes, random noise, drift, and stiction effects. Table \ref{tab:faults} lists each fault along with its identifier, disturbance type, and the affected variable or mechanism. Faults 16-20 were originally labeled as unknown in the TEP benchmark. Bathelt et al. \cite{Bathelt2015} later disclosed the mechanisms for Faults 16-19 as part of their benchmark revision. Fault 20, however, remained undocumented. We addressed it in our previous work~\cite{Luna2025}, providing an explanation of the disturbance and its impact on process behavior.

\begin{table}[t]
\centering
\caption{Summary of fault scenarios in the TEP benchmark.}
\label{tab:faults}
\scriptsize
\begin{tabular}{ccp{0.6\columnwidth}}
\toprule
\textbf{Fault} & \textbf{Type} & \textbf{Disturbed Value} \\
\midrule
1 & Step & A/C-ratio of stream 4, B composition constant \\
2 & Step & B composition of stream 4, A/C-ratio constant \\
3 & Step & D feed (stream 2) temperature \\
4 & Step & Cooling water inlet temperature of reactor \\
5 & Step & Cooling water inlet temperature of separator \\
6 & Step & A feed loss (stream 1) \\
7 & Step & C header pressure loss (stream 4) \\
8 & Random & A/B/C composition of stream 4 \\
9 & Random & D feed (stream 2) temperature \\
10 & Random & C feed (stream 4) temperature \\
11 & Random & Cooling water inlet temperature of reactor \\
12 & Random & Cooling water inlet temperature of separator \\
13 & Drift & Reaction kinetics \\
14 & Stiction & Cooling water outlet valve of reactor \\
15 & Stiction & Cooling water outlet valve of separator \\
16 & Random & Deviations in heat transfer within stripper \\
17 & Random & Deviations in heat transfer within reactor \\
18 & Random & Deviations in heat transfer within condenser \\
19 & Stiction & Recycle valve of compressor, underflow separator (stream 10), underflow stripper (stream 11), and steam valve stripper \\
20 & Random & Deviations in flow rate (stream 7) \\
21 & Constant position & Valve (Stream 4) \\
\bottomrule
\end{tabular}
\end{table}

\begin{table}[t]
\centering
\caption{Subset sizes (prior to train/validation split) for training AE variants.}
\label{tab:subset_sizes}
\scriptsize
\begin{tabular}{l l}
\toprule
\textbf{Description} & \textbf{Subset Sizes} \\
\midrule
Number of samples & 
625,\ 1250,\ 1875,\ 3125,\ 5000,\ 8125,\ 13750, \\
& 23125,\ 38125,\ 64375,\ 107500,\ 180000,\ 250000 \\
\bottomrule
\end{tabular}
\end{table}

\subsection{Data Preprocessing}
\label{sec:data_preprocessing}
We selected a subset of 33 variables from the TEP benchmark by excluding $\mathrm{XMEAS}(23)$ to $\mathrm{XMEAS}(41)$, which correspond to analyzer measurements in the reactor feed, purge stream, and final product stream. This exclusion, consistent with the methodology proposed by Cacciarelli et al.~\cite{Cacciarelli2022}, reflects the limited availability of analyzer data in real-time industrial operations. As in prior work, each sample represents a single time step, with no temporal context or windowing applied.

To evaluate how detection performance scales with training data availability, we constructed 13 fault-free subsets ranging from 625 to 250,000 samples (see Table~\ref{tab:subset_sizes}). Each subset was split into 80\% training and 20\% validation sets. To preserve temporal structure, the data was split at the simulation level: full simulation runs were randomly selected for training and validation, with the final simulation sliced if necessary to achieve the desired number of samples.

A z-score normalization scaler was fitted to the training set and applied consistently to both partitions. The scaler was saved and reused during inference to ensure consistent preprocessing. For reproducibility, fixed random seeds were used when generating data splits. These seeds were shared across model variants during training runs to ensure matched conditions. Further implementation details are provided in Section~\ref{sec:training}.

\begin{table}[t]
\centering
\caption{Overview of monitoring model settings.}
\label{tab:model_config_summary}
\scriptsize
\begin{tabular}{lll}
\toprule
\textbf{Model} & \textbf{Attribute} & \textbf{Value} \\
\midrule

\multirow{6}{*}{OAE} 
  & Layer sizes & [33, 85, 25, 85, 33] \\
  & Function parameterization & Multilayer perceptron \\
  & Activation function & ReLU \\
  & Regularization & Orthogonality and weight decay  \\
  & No. of Parameters & 10,088 \\

\midrule
\multirow{5}{*}{EfficientKAN-AE} 
  & Layer sizes & [33, 25, 33] \\
  & Function parameterization & SiLU + B-spline expansion \\
  & No. of basis functions & 6 \\
  & Regularization & L1, entropy-based and weight decay  \\
  & No. of parameters & 11,550 \\

\midrule
\multirow{5}{*}{FastKAN-AE} 
  & Layer sizes & [33, 25, 33] \\
  & Function parameterization & SiLU + Gaussian RBF expansion \\
  & No. of basis functions & 5 \\
  & Regularization & Weight decay \\
  & No. of parameters & 10,074 \\

\midrule
\multirow{5}{*}{FourierKAN-AE} 
  & Layer sizes & [33, 25, 33] \\
  & Function parameterization & Fourier expansion \\
  & No. of basis functions & 7 \\
  & Regularization & Weight decay \\
  & No. of parameters & 9,958 \\

\midrule
\multirow{4}{*}{WavKAN-AE} 
  & Layer sizes & [33, 25, 33] \\
  & Function parameterization & Single DoG wavelet \\
  & Regularization & Weight decay \\
  & No. of parameters & 6,716 \\

\bottomrule
\end{tabular}
\end{table}

\subsection{Process Monitoring Models}
All models follow a common autoencoder structure, differing only in how the encoder and decoder mappings are parameterized. As in the benchmarking protocol of Cacciarelli and Kulahci \cite{Cacciarelli2022}, each model processes input samples with 33 features (see Section \ref{sec:TEP}) and compresses them to a 25-dimensional latent space. Models are trained to minimize the mean squared reconstruction error (Eq. \ref{eq:loss_ae}), with additional regularization terms applied as dictated by each architecture. Specifically, the OAE employs orthogonality regularization (Eq. \ref{eq:loss_oae}), while EfficientKAN-AE introduces both L1 and entropy-based penalties (Eqs. \ref{eq:l1_activity}, \ref{eq:entropy_activity}). In addition, decoupled weight decay \cite{loshchilov2019} is applied across all models as a shared regularization strategy.      

Table~\ref{tab:model_config_summary} summarizes key architectural and regularization attributes of the evaluated models. 

\subsection{Training Procedure}
\label{sec:training}
Each AE variant was trained independently on each subset of fault-free data (see Section \ref{sec:data_preprocessing}), with 30 random seeds per setting to account for stochastic variability from weight initialization and mini-batch sampling. This setup yields 390 independent training runs per model type. 

Training was performed using the AdamW optimizer with a mini-batch size $B = 256$ and a maximum of $E_{\max} = 600$ epochs. Mixed-precision training was enabled via PyTorch \texttt{AMP} module to accelerate training and reduce memory usage. A \texttt{ReduceLROnPlateau} schedule reduced the learning rate after 5 validation epochs without improvement, and training was terminated early if no improvement was observed for 15 consecutive epochs. These settings were held constant across all AE variants.

The initial learning rate $\eta_0$, weight decay, and scheduler reduction factor were optimized for each AE variant using the Tree-structured Parzen Estimator (TPE) algorithm \cite{bergstra2011}, implemented via the Optuna framework \cite{akiba2019}. For the OAE and EfficientKAN-AE models, additional tuning was performed for the orthogonality and sparsity regularization coefficients, respectively. Hyperparameter tuning was conducted on a fixed subset of 625 fault-free samples, split 80/20 into training and validation sets. Since fault labels are not used during training, the reconstruction loss on the validation set was used as an unsupervised proxy for detection performance. The final hyperparameter values for each model are listed in Table~\ref{tab:training_config_summary}.

\subsection{Fault Detection}
Autoencoders are widely used for fault detection in industrial systems, particularly in settings where labeled fault data are scarce or unavailable. Trained solely on normal process data, autoencoders learn compact latent representations that allow accurate reconstruction of fault-free observations. Faults typically disrupt these patterns, leading to elevated reconstruction errors that indicate abnormal behavior.

Two monitoring statistics are commonly used in AE-based fault detection: the squared prediction error (SPE), also known as the \( Q \)-statistic, and Hotelling’s \( T^2 \) statistic. The former measures the difference between an input and its reconstruction, while the latter measures the deviation of a sample’s latent representation from the center of the fault-free data distribution in the latent space. Although both statistics can be used jointly to improve detection sensitivity and robustness, this study focuses exclusively on the SPE.

For a given input sample \( x_i \), the SPE is defined as:

\begin{equation}
Q_i = \|x_i - \hat{x}_i\|^2_2,
\label{eq:Q_stat}
\end{equation}
\noindent
where \( \hat{x}_i \) denotes the reconstruction of \( x_i \). Elevated values of \( Q_i \) indicate that the model is unable to reconstruct the input, suggesting faulty behavior. 

To determine whether a given \( Q_i \) indicates a fault, a detection threshold \( Q_{\text{lim}} \) is estimated using kernel density estimation (KDE). This nonparametric method avoids assumptions about the underlying distribution of SPE values, offering flexibility in modeling nonlinear reconstruction behavior. The estimated probability density function $\hat{P}(Q)$ is given by:

\begin{equation}
\hat{P}(Q) = \frac{1}{n h} \sum_{i=1}^n K\left( \frac{Q - Q_i}{h} \right),
\label{eq:KDE}
\end{equation}
\noindent
where $Q_i$ represents the observed SPE values under normal conditions, $K(\cdot)$ is the Gaussian kernel function, and $h$ is the bandwidth parameter. The bandwidth is determined via Scott’s Rule \cite{scott1992}:

\begin{equation}
h = \sigma n^{-1/5},
\end{equation}
\noindent
where \(\sigma\) is the standard deviation of $Q_i$ values and $n$ is the number of training samples.

The detection threshold \( Q_{\text{lim}} \) is set to the ($1 - \alpha$)-quantile of the estimated density:

\begin{equation}
Q_{\text{lim}} = \inf\left\{x \mid \int_{-\infty}^x \hat{P}(Q) \, dQ \geq 1 - \alpha \right\},
\label{eq:Q_limit}
\end{equation}
\noindent
where $\alpha$ is a specified significance level. In this study, we set $\alpha = 0.05$, corresponding to an expected false alarm rate of approximately 5\% under normal operating conditions.

\begin{table}[t]
\centering
\caption{Optimized training hyperparameters per AE variant (via TPE).}
\label{tab:training_config_summary}
\scriptsize
\begin{tabular}{lll}
\toprule
\textbf{Model} & \textbf{Attribute} & \textbf{Value} \\
\midrule

\multirow{4}{*}{OAE} 
  & Initial learning rate & $1.00 \times 10^{-3}$ \\
  & Weight decay & $1.00 \times 10^{-2}$ \\
  & Scheduler factor& 0.20 \\
  & Orthogonality coefficient & $1.00$  \\

\midrule
\multirow{5}{*}{EfficientKAN-AE} 
  & Initial learning rate & $4.38 \times 10^{-3}$ \\
  & Weight decay & $2.00 \times 10^{-2}$ \\
  & Scheduler factor& 0.96 \\
  & L1 coefficient & $1.93 \times 10^{-4}$  \\
  & Entropy coefficient & $7.73 \times 10^{-4}$  \\

\midrule
\multirow{3}{*}{FastKAN-AE} 
  & Initial learning rate & $1.92 \times 10^{-3}$ \\
  & Weight decay & $9.60 \times 10^{-3}$ \\
  & Scheduler factor& 0.93 \\
  
\midrule
\multirow{3}{*}{FourierKAN-AE} 
  & Initial learning rate & $3.63 \times 10^{-3}$ \\
  & Weight decay & $5.38 \times 10^{-3}$ \\
  & Scheduler factor& 0.98 \\

\midrule
\multirow{3}{*}{WavKAN-AE} 
  & Initial learning rate & $4.99 \times 10^{-3}$ \\
  & Weight decay & $7.60 \times 10^{-3}$ \\
  & Scheduler factor& 0.95 \\

\bottomrule
\end{tabular}
\end{table}

\subsection{Performance Evaluation}
Each trained model is evaluated on a fixed test dataset containing both fault-free and faulty samples. To determine whether a test sample is normal or faulty, we use the previously defined threshold $Q_{\text{lim}}$. Samples with \( Q_i > Q_{\text{lim}} \) are flagged as faulty.

Fault detection performance is quantified using two standard metrics: Fault Detection Rate (FDR) and False Alarm Rate (FAR), defined as:

\begin{equation}
\text{FDR} = \frac{\text{TP}}{\text{TP} + \text{FN}},
\end{equation}

\begin{equation}
\text{FAR} = \frac{\text{FP}}{\text{FP} + \text{TN}},
\end{equation}
\noindent
where TP, FP, FN, and TN denote the number of true positives, false positives, false negatives, and true negatives, respectively. FDR measures the proportion of actual faults correctly detected, while FAR captures the frequency of erroneous alarms. Together, these metrics characterize the trade-off between sensitivity and specificity. Therefore, robust fault detection requires models that consistently achieve high FDR while maintaining a low FAR.

\subsection{Bayesian Signed-Rank Test for Model Comparison}
\label{sec:bayesian-signrank1}
While FDR summarizes average detection performance, it does not indicate whether observed differences in performance between models are statistically significant. We adopt the Bayesian signed-rank test introduced by Benavoli et al.~\cite{Benavoli2014} to assess whether one model consistently outperforms another across multiple faults. This method estimates the posterior probability that one model is practically superior to another, explicitly accounting for sampling uncertainty and defining a region of practical equivalence (ROPE) where small differences are considered negligible. In this study, we set the ROPE radius to \( r = 0.01 \), corresponding to a $\pm1\%$ change in absolute FDR.

For each fault \( i \in \{1, \dots, q\} \), we calculate the difference in mean FDR across 30 training seeds:
\begin{equation}
\Delta_i = \mu_{B,i} - \mu_{A,i},
\end{equation}
where \( \mu_{A,i} \) and \( \mu_{B,i} \) denote the average FDRs for models A and B, respectively. The set \( \{\Delta_1, \dots, \Delta_q\} \) is treated as an i.i.d. sample from an unknown distribution over model performance differences. To capture this uncertainty, we place a nonparametric Dirichlet Process (DP) prior over the distribution:
\begin{equation}
G \sim \mathrm{DP}(s, \delta_0),
\end{equation}
where $s$ is the concentration parameter and \( \delta_0 \) is a Dirac measure centered at zero. This prior reflects a conservative belief that the models are practically equivalent unless strong evidence suggests otherwise.

The posterior predictive distribution over future performance differences takes the form:
\begin{align}
p(z) &= w_0 \delta_0(z) + \sum_{i=1}^q w_i \delta_{\Delta_i}(z), \label{eq:posterior-mixture} \\
(w_0, w_1, \dots, w_q) &\sim \mathrm{Dir}(s, 1, \dots, 1). \label{eq:dirichlet}
\end{align}

To determine which model is favored, we estimate posterior probabilities via Monte Carlo sampling. Let \( z_0 = 0 \) denote the prior pseudo-observation and \( z_1, \dots, z_q \) the observed differences across faults. At each iteration, a weight vector \( (w_0, w_1, \dots, w_q) \) is drawn from a Dirichlet distribution with parameters \( (s, 1, \dots, 1) \). These weights define a posterior sample over the discrete distribution of differences.

For each sampled belief, we compute the probabilities that the sum of two independent draws falls into one of three regions:

\begin{align}
\theta_l &= \sum_{i,j} w_i w_j \cdot \mathbb{I}(z_i + z_j < -2r), \\
\theta_e &= \sum_{i,j} w_i w_j \cdot \mathbb{I}(|z_i + z_j| \leq 2r), \\
\theta_r &= \sum_{i,j} w_i w_j \cdot \mathbb{I}(z_i + z_j > 2r),
\end{align}

where \( \mathbb{I}(\cdot) \) is the indicator function. The quantities $\theta_l,\theta_e,\theta_r$ represent the probability mass in each outcome region for a single posterior sample. 

Repeating this process \( N \) times yields a collection of triplets \( (\theta_l^{(1)}, \theta_e^{(1)}, \theta_r^{(1)}), \dots, (\theta_l^{(N)}, \theta_e^{(N)}, \theta_r^{(N)}) \), which define an empirical approximation of the posterior distribution over region probabilities. Following Benavoli et al.~\cite{Benavoli2017}, we summarize the overall belief in each outcome by computing the proportion of samples in which that region has the highest probability. 

For example, the probability that model A is practically superior to model B is given by:
\begin{equation}
P_{\text{left}} = \frac{1}{N} \sum_{i=1}^N \mathbb{I}\left(\theta_l^{(i)} > \max\{\theta_e^{(i)}, \theta_r^{(i)}\}\right),
\end{equation}
with analogous definitions for \( P_{\text{rope}} \) and \( P_{\text{right}} \). These quantities summarize how often each outcome dominates in the posterior, offering a probabilistic account of relative model performance.

\begin{figure*}[t]
    \centering
    \includegraphics[width=0.72\textwidth]{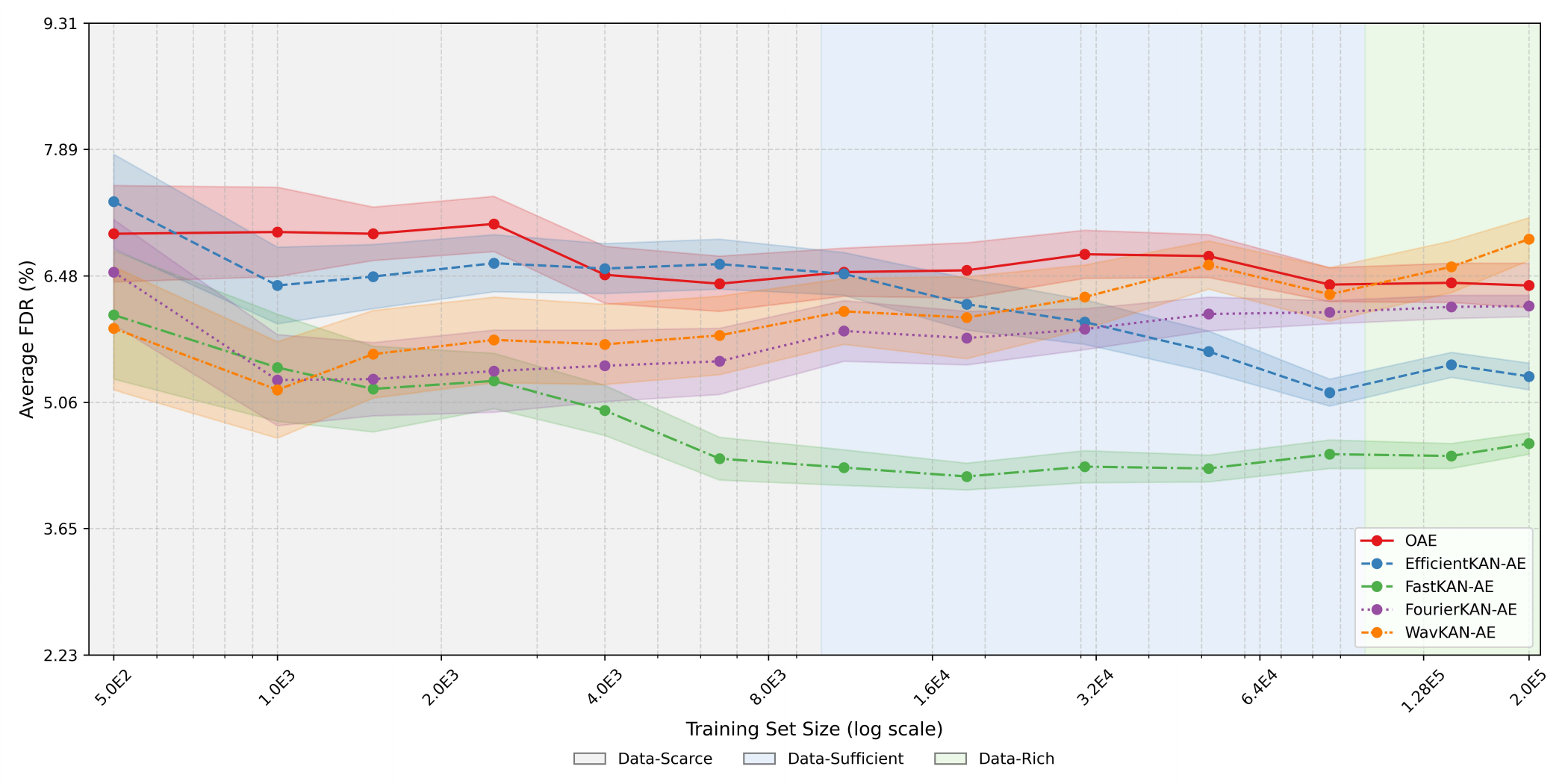}
    \caption{Fault Detection Rate (FDR) across different training sample sizes for controllable faults.\label{fig:controllable_fdr}}
\end{figure*}

\begin{figure*}[t]
    \centering
    \includegraphics[width=0.72\textwidth]{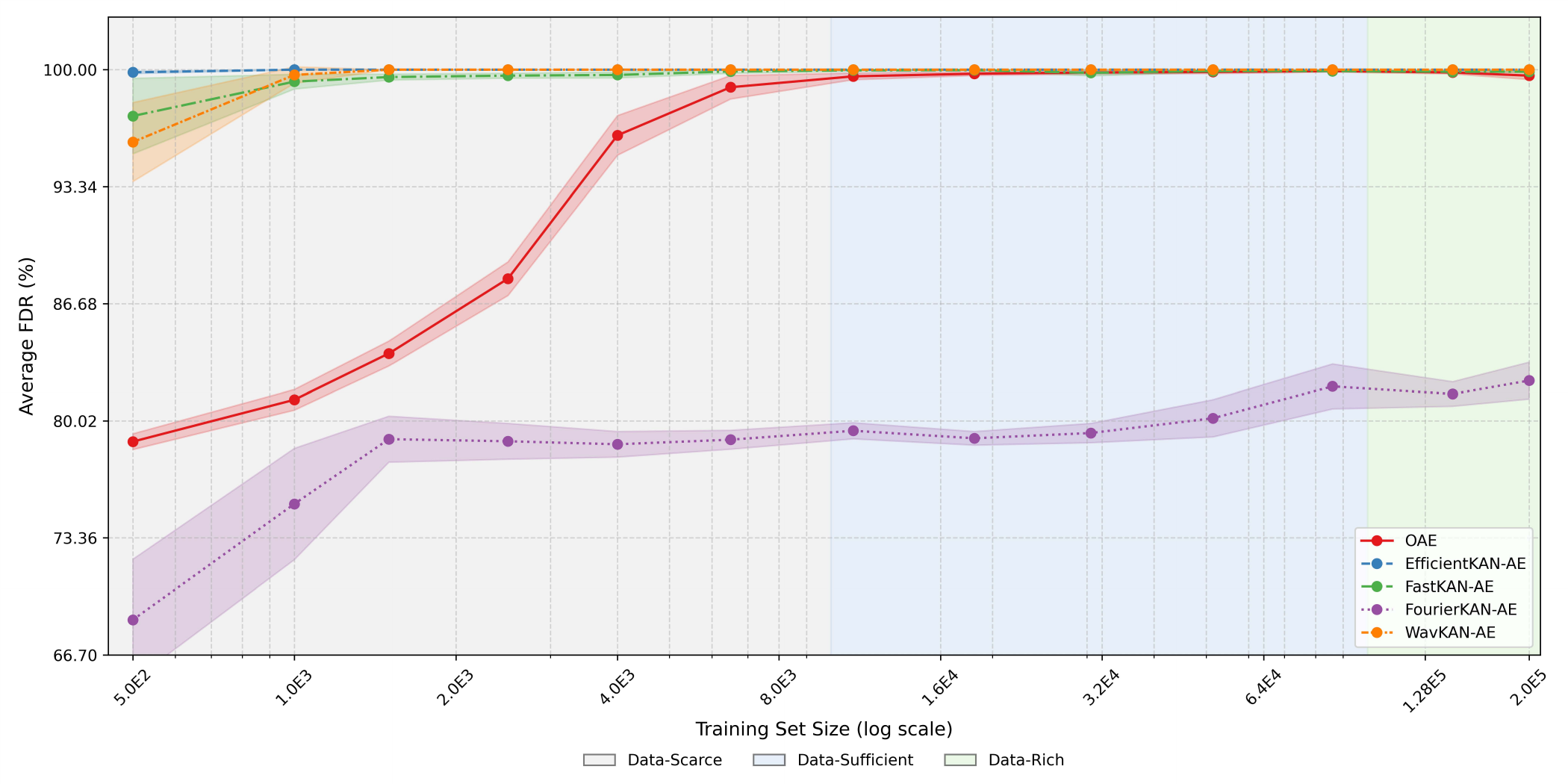}
    \caption{Fault Detection Rate (FDR) across different training sample sizes for back-to-control faults.\label{fig:back-to-control_fdr}}
\end{figure*}

\begin{figure*}[h!]
    \centering
    \includegraphics[width=0.72\textwidth]{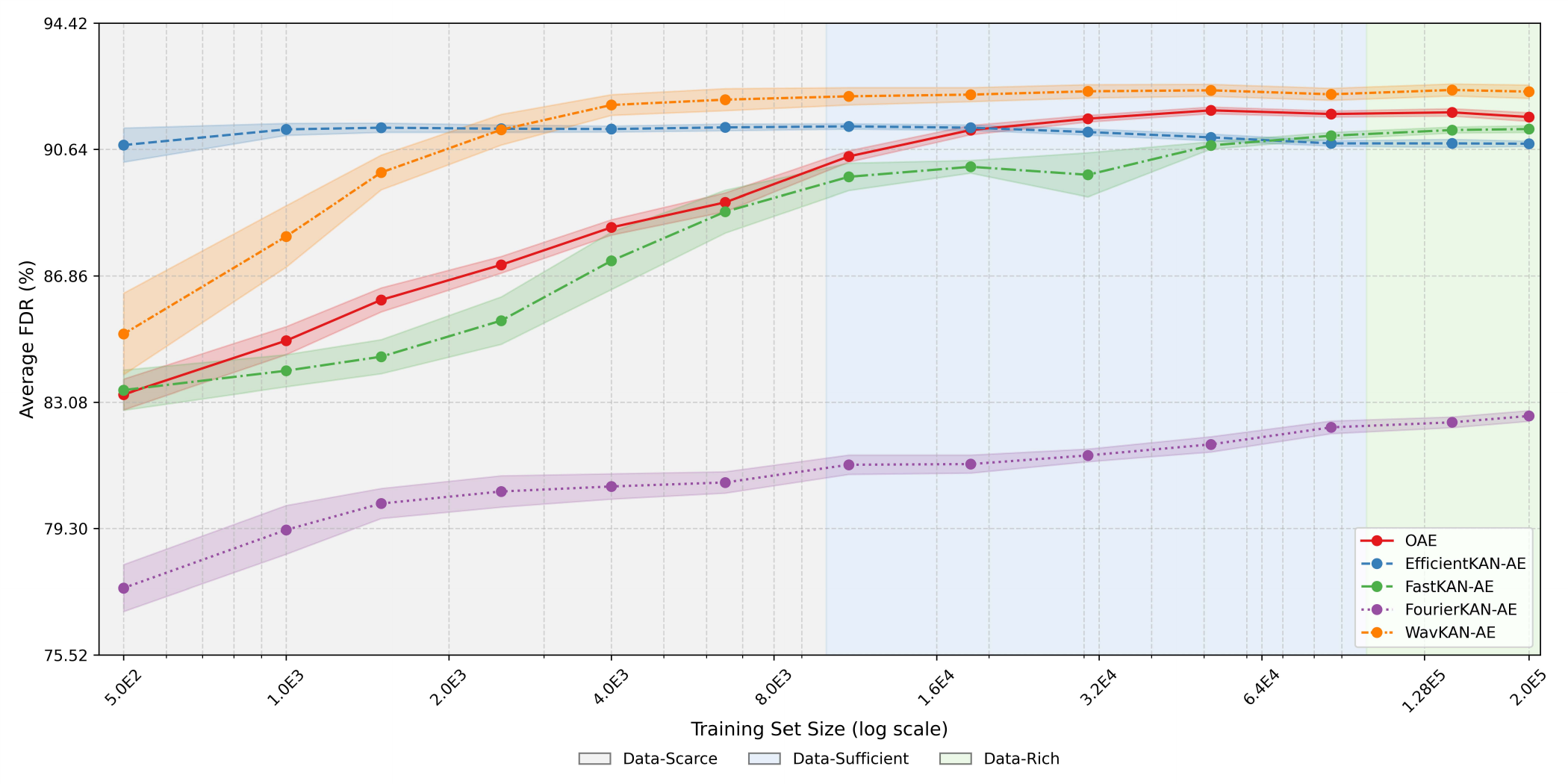}
    \caption{Fault Detection Rate (FDR) across different training sample sizes for uncontrollable faults.\label{fig:uncontrollable_fdr}}
\end{figure*}

\begin{figure*}[t]
    \centering

    \begin{subfigure}[t]{0.31\textwidth}
        \includegraphics[width=\linewidth]{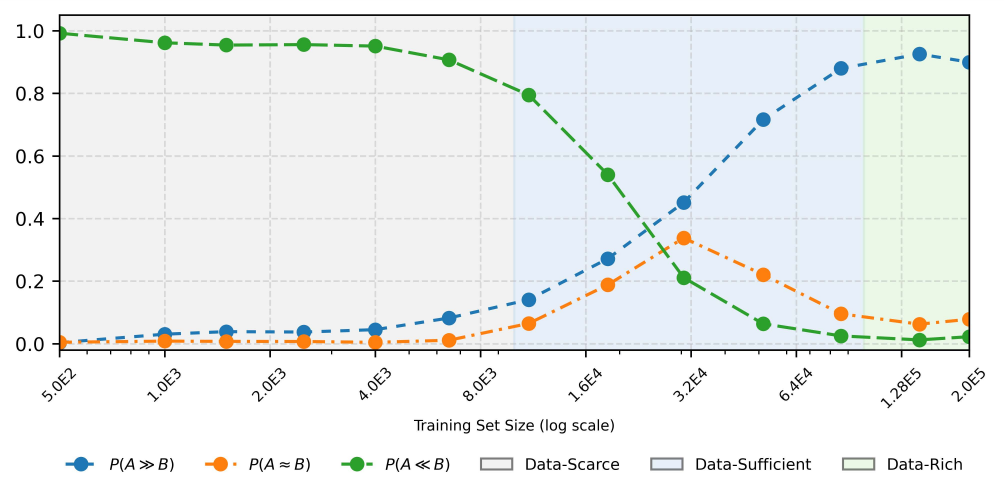}
        \subcaption{A = OAE, B = EfficientKAN-AE}
        \label{fig:OAEvsEffKAN} 
    \end{subfigure}
    \hfill
    \begin{subfigure}[t]{0.31\textwidth}
        \includegraphics[width=\linewidth]{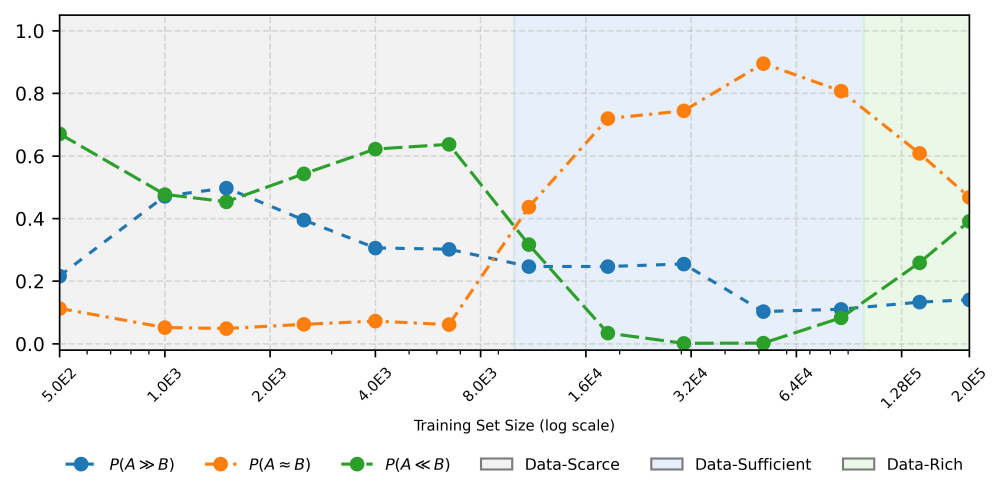}
        \subcaption{A = OAE, B = FastKAN-AE}
        \label{fig:OAEvsFastKAN}
    \end{subfigure}
    \hfill
    \begin{subfigure}[t]{0.31\textwidth}
        \includegraphics[width=\linewidth]{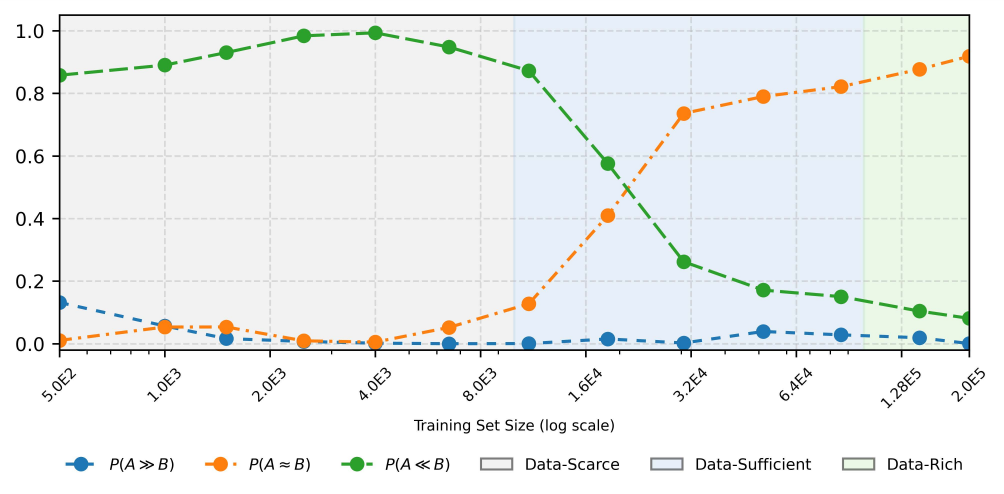}
        \subcaption{A = OAE, B = WavKAN-AE}
        \label{fig:OAEvsWavKAN}
    \end{subfigure}

    \vspace{1em}

    \begin{subfigure}[t]{0.31\textwidth}
        \includegraphics[width=\linewidth]{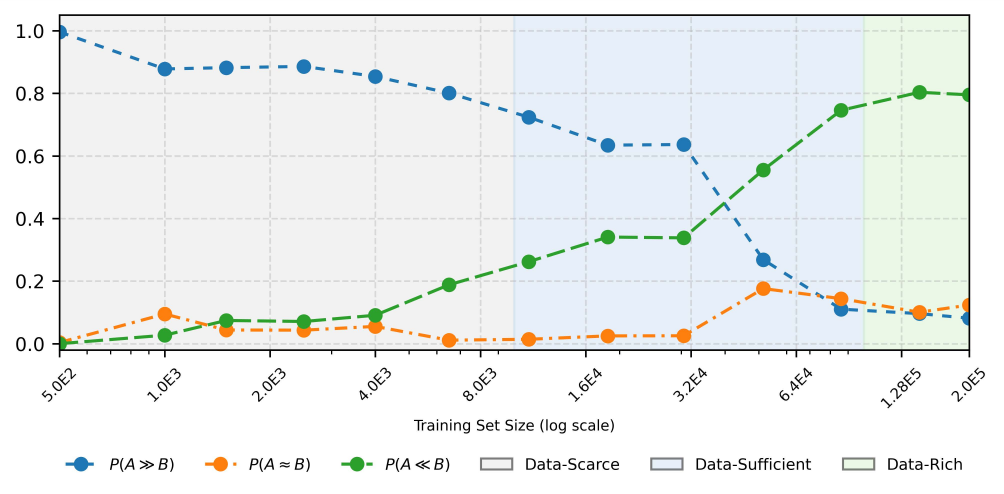}
        \subcaption{A = EfficientKAN-AE, B = FastKAN-AE}
        \label{fig:EffKANvsFastKAN}
    \end{subfigure}
    \hfill
    \begin{subfigure}[t]{0.31\textwidth}
        \includegraphics[width=\linewidth]{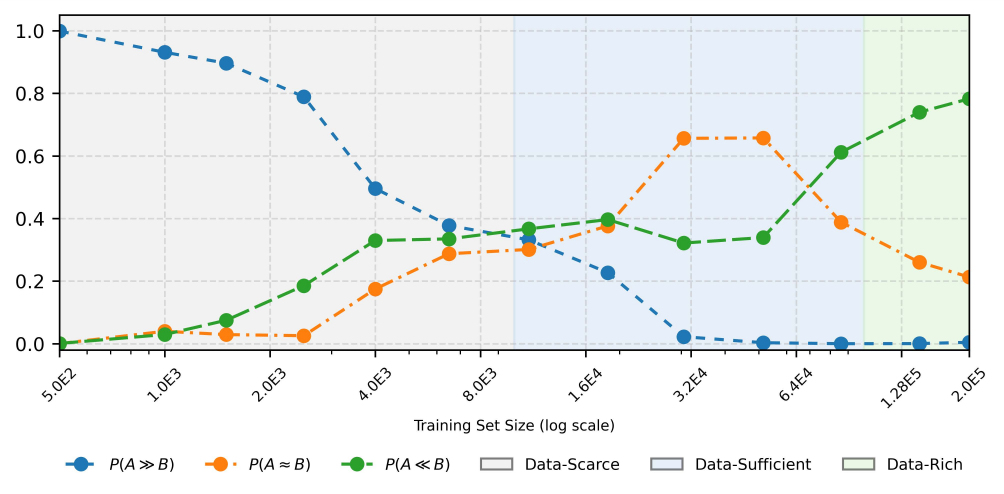}
        \subcaption{A = EfficientKAN-AE, B = WavKAN-AE}
        \label{fig:EffKANvsWavKAN}
    \end{subfigure}
    \hfill
    \begin{subfigure}[t]{0.31\textwidth}
        \includegraphics[width=\linewidth]{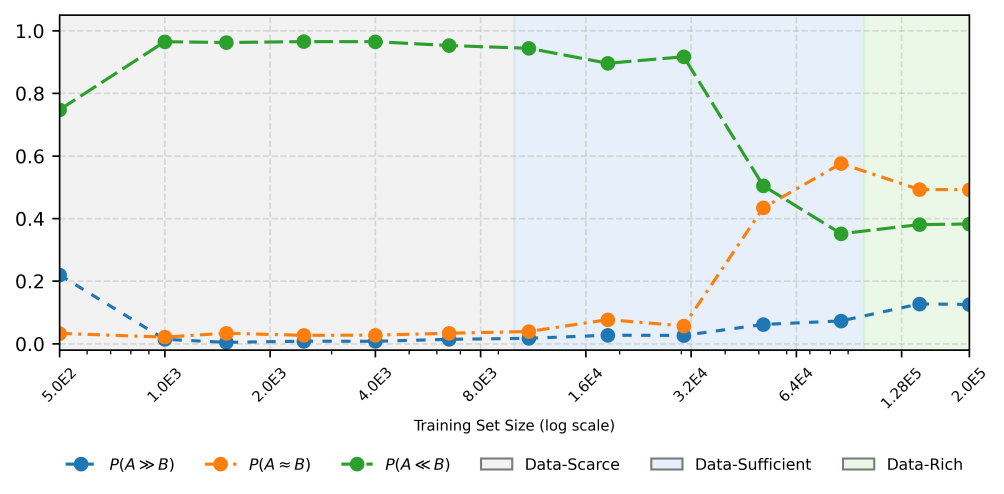}
        \subcaption{A = FastKAN-AE, B = WavKAN-AE}
        \label{fig:FastKANvsWavKAN}
    \end{subfigure}

    \caption{Evolution of Bayesian signed-rank probabilities across training regimes. Each subfigure compares two autoencoder variants. The three shaded zones denote data-scarce, data-sufficient, and data-rich training regimes, respectively.}
    \label{fig:bayesian-profiles}
\end{figure*}

\section{Results and Discussion}
\label{sec:results and discussion}
\subsection{Performance Scaling with Training Set Size}
\label{sec:scaling_fdr}
To evaluate model performance across fault types, we adopt the classification proposed by Sun et al. \cite{Sun2020}, which groups faults into three categories: controllable, back-to-control, and uncontrollable (see Section \ref{sec:TEP}). High FDRs are desirable for back-to-control and uncontrollable faults, as they reflect timely and effective detection. Conversely, elevated FDRs for controllable faults may reflect excessive model sensitivity and lead to unnecessary process interventions. An effective monitoring system should therefore achieve high FDRs for back-to-control and uncontrollable faults while keeping the FDR near the nominal false alarm level (i.e., $\text{FDR} \approx 0.05$) for controllable ones. 

Training set sizes (denoted as $n_{\text{train}}$) are grouped into three regimes: data-scarce ($n_{\text{train}} < 10^4$), data-sufficient ($10^4 \leq n_{\text{train}} \leq 10^5$), and data-rich ($n_{\text{train}} > 10^5$). For each model, FDRs are first averaged across faults within each category, then across training runs. Figures~\ref{fig:controllable_fdr}–\ref{fig:uncontrollable_fdr} show these results with 95\% confidence intervals to reflect training variability.

\subsubsection{Controllable Faults}
Across all training set sizes, the average FDR on controllable faults remains within a narrow range of 6–7\% for most models (Figure~\ref{fig:controllable_fdr}). This stability suggests that increasing the amount of fault-free data has minimal impact on detection sensitivity in this category, likely due to the low-amplitude and self-correcting nature of these faults. OAE exhibits nearly constant performance, while EfficientKAN shows a slight decrease in FDR with larger datasets. FastKAN-AE demonstrates the most notable improvement, reaching an average FDR of 4.6\% as $n_{\text{train}}$ increases. In contrast, both WavKAN-AE and FourierKAN-AE display a marginal rise in FDR, suggesting a mild over-sensitivity to normal fluctuations that may be misinterpreted as fault signatures.

\subsubsection{Back-to-Control Faults}
For back-to-control faults, most models' detection performance improved as $n_{\text{train}}$ increased (Figure~\ref{fig:back-to-control_fdr}). EfficientKAN-AE and WavKAN-AE achieved near-perfect FDRs with as few as 1,000 samples and maintained this performance at larger scales. Their early saturation underscores strong data efficiency in capturing persistent deviations in manipulated variables. FastKAN-AE followed a similar trajectory but required more data to reach peak performance, indicating reduced data efficiency. OAE also improved steadily, eventually achieving high FDRs in the data-sufficient regime, though its improvement was slower compared to the KAN-based models. This indicates that even though OAE eventually reaches high FDR, KAN variants achieve comparable detection with far fewer samples. In contrast, FourierKAN-AE consistently underperformed, with FDR plateauing well below the levels achieved by other models. This behavior likely reflects the limited expressiveness of its low-order global basis functions, which may struggle to capture the systematic deviations characteristic of this fault category.

\subsubsection{Uncontrollable Faults}
In the case of uncontrollable faults, detection performance improved with $n_{\text{train}}$ across all models, though the rate and extent of improvement varied (Figure~\ref{fig:uncontrollable_fdr}). EfficientKAN-AE started with the highest FDR at $n_{\text{train}}=500$ and maintained stable performance throughout, with only a slight dip observed in the data-sufficient regime, potentially due to over-regularization effects. In contrast, WavKAN-AE began with lower FDR but improved rapidly, remarkably achieving over 92\% FDR with only 2,500 training samples. This rapid convergence suggests that its wavelet-based parameterization is particularly effective for detecting persistent, uncorrected deviations even under limited data. FastKAN-AE displayed a delayed performance response, with higher variability across seeds and only exhibiting strong performance in the data-rich regime. OAE followed a more gradual improvement curve, eventually achieving competitive performance in the data-sufficient regime, although it did not surpass WavKAN-AE even at larger training set sizes. FourierKAN-AE consistently lagged behind, reinforcing the limitations of its global basis representation in capturing fault-specific deviations.

\subsection{Bayesian Comparison across Training Set Sizes}
\label{sec:bayesian-signrank2}
To determine whether observed differences in detection performance are statistically meaningful, we apply the Bayesian signed-rank test described in Section~\ref{sec:bayesian-signrank1}. This analysis focuses on a subset of challenging faults, as these cases are more likely to reveal performance differences among models. Specifically, we exclude faults where PCA achieves near-perfect detection (FDR $> 95\%$, as reported in~\cite{Sun2020}), as well as controllable faults, where elevated FDRs may indicate over-sensitivity rather than genuine detection capability. The resulting evaluation set comprises nine faults: Fault 5, 10, 11, 16--21, each treated as an independent comparison unit.

For each model pair, we compute the posterior probabilities \( P_{\text{left}} \), \( P_{\text{rope}} \), and \( P_{\text{right}} \), which quantify the belief that model A is practically superior to model B, that their performance is equivalent within a ROPE, or that model B is better. These probabilities are tracked across training set sizes to capture how relative performance evolves as more data become available. The resulting trends complement the mean FDR profiles discussed in Section \ref{sec:scaling_fdr}, providing a probabilistic basis for model selection as a function of $n_{\text{train}}$.

FourierKAN-AE is omitted from this analysis due to its consistently poor performance across fault types and training set sizes. Including it in pairwise tests would not yield informative distinctions, since posterior comparisons involving clearly dominated models tend to collapse onto the extremes (e.g., $P_{\text{right}} \approx 1$), offering little inferential value. 

\subsubsection{Orthogonal Autoencoder vs. KAN-based Autoencoders}
As shown in Fig. \ref{fig:OAEvsEffKAN} EfficientKAN demonstrates a clear advantage in the data-scarce regime. For $n_{\text{train}} \leq 6,500$, the posterior mass lies overwhelmingly in \(P_{\text{right}}\) (\(>0.90\)), peaking at 0.992 for $n_{\text{train}}=500$. This provides strong evidence that EfficientKAN-AE outperforms OAE when data are limited. However, this advantage diminishes steadily as $n_{\text{train}}$ increases. Between 11,000 and 30,000 training samples, the posterior probability begins to shift toward the ROPE, and eventually to \(P_{\text{left}}\), suggesting a reversal in relative performance. At $n_{\text{train}}=51,500$, OAE is favored with \(P_{\text{left}}=0.72\), and this belief strengthens to \(P_{\text{left}}=0.93\) by 144,000 training samples. These results highlight EfficientKAN-AE's strong data efficiency, but also its declining competitiveness as more training data become available.

FastKAN-AE exhibits a more ambiguous performance profile. At small $n_{\text{train}}$ (e.g., 4,000-6,500), it holds a slight advantage, with \(P_{\text{right}}\) peaking at 0.64. However, this advantage is never decisive, and the posterior mass rapidly shifts toward the ROPE as the $n_{\text{train}}$ increases. Beyond $n_{\text{train}}=18,500$, practical equivalence dominates the posterior (\(P_{\text{rope}} > 0.70\)), reaching as high as 0.90 at $n_{\text{train}}=51,500$. Notably, FastKAN is never overtaken by OAE, as \(P_{\text{left}}\) remains consistently low. However, it also fails to establish a clear advantage, with \(P_{\text{right}}\) never surpassing 0.70 (see Fig. \ref{fig:OAEvsFastKAN}). This pattern suggests that FastKAN-AE approaches a level of performance similar to OAE across most data regimes, providing only marginal gains in fault detection performance over the baseline.

WavKAN-AE delivers the most robust early-stage performance. For $n_{\text{train}} \leq 6,500$, \(P_{\text{right}}\) consistently exceeds 0.94, peaking at 0.993 for $n_{\text{train}}=4,000$. As the $n_{\text{train}}$ becomes larger, the posterior mass gradually shifts toward the ROPE region, indicating a tendency toward practical equivalence rather than a reversal in relative performance (see Fig. \ref{fig:OAEvsWavKAN}). By $n_{\text{train}}=30,500$, ROPE accounts for 74\% of the posterior mass, rising to 92\% by $n_{\text{train}}=200,000$. Importantly, \(P_{\text{left}}\) remains near zero throughout, meaning OAE never overtakes WavKAN-AE. This underscores WavKAN-AE’s robustness in data-scarce regimes and its ability to maintain competitive performance as more training data become available, without the degradation observed in EfficientKAN-AE.

Collectively, these results reveal distinct generalization behaviors across KAN-based architectures. EfficientKAN-AE exhibits strong performance in low-data regimes but is ultimately outperformed as training set size increases. FastKAN-AE yields modest gains early on and tends toward practical equivalence, without establishing dominance. WavKAN-AE delivers the strongest performance gains in low-data regimes and continues to perform competitively as the training set size increases. These findings underscore the importance of aligning model choice with data availability: EfficientKAN-AE is preferable when training data are scarce, WavKAN-AE offers strong and stable performance across regimes, and OAE remains a competitive option in data-rich scenarios.

\subsubsection{Pairwise Comparisons among KAN-based Autoencoders}
In the data-scarce regime, EfficientKAN-AE consistently outperforms the other KAN-based variants by a substantial margin. Posterior probability mass favoring EfficientKAN-AE exceeds 0.85 in all pairwise comparisons, with \(P_{\text{right}}\) reaching 0.996 versus FastKAN-AE and 0.999 versus WavKAN-AE, strongly supporting EfficientKAN-AE’s superiority (Fig.~\ref{fig:EffKANvsFastKAN}-\ref{fig:EffKANvsWavKAN}). Within the same regime, WavKAN-AE maintains a moderate but consistent advantage over FastKAN-AE, with posterior mass exceeding 0.74 (Fig.~\ref{fig:FastKANvsWavKAN}), though this margin is notably smaller than that of EfficientKAN-AE.

As the training set enters the data-sufficient regime, WavKAN-AE emerges as the most robust model. It decisively outperforms FastKAN-AE throughout this range, with \(P_{\text{right}} > 0.90\) across all relevant comparisons. In comparisons with EfficientKAN-AE, the posterior distribution shifts more gradually. Initially, a broad region of practical equivalence dominates, but belief in WavKAN-AE’s superiority strengthens with additional data and eventually overtakes EfficientKAN-AE midway through the regime.

In the data-rich regime, both WavKAN-AE and FastKAN-AE surpass EfficientKAN-AE. Posterior mass favoring FastKAN-AE over EfficientKAN-AE increases steadily, reaching 0.80 at the upper end of the regime. WavKAN-AE also continues to outperform EfficientKAN-AE, reaching \(P_{\text{right}} = 0.78\). However, FastKAN-AE never overtakes WavKAN-AE at any point, as its posterior advantage, though growing marginally with scale, remains below 0.40 throughout. In large-data comparisons between WavKAN-AE and FastKAN-AE, posterior mass is concentrated within the ROPE, suggesting practical equivalence. 

\subsection{Model Behavior in the Data-Scarce Limit (500 Samples)}
In many process monitoring applications, access to extensive fault-free historical data may be limited due to operational variability, economic constraints, or short observation windows. As such, evaluating model performance under extreme data scarcity is critical for assessing robustness and real-world applicability. This section focuses on the training done with 500 samples, which represents the most constrained setting considered in this study, to examine how well each model generalizes from minimal input data. Detection behavior in this limit offers insight into architectural biases and data efficiency, particularly the role of edge function parameterization. A detailed per-fault comparison is used to assess each model’s early-stage detection capability.

\subsubsection{Fault Detection Rates}
All models exhibit relatively low FDRs on controllable faults (3, 9, and 15), suggesting limited response to self-correcting process variations. WavKAN-AE achieves the lowest values on Faults 3 and 9, reflecting a stronger tendency to suppress false positives, likely due to its localized wavelet basis. However, FDRs on Fault 15 are slightly elevated across all models (7–9\%), exceeding the nominal false alarm rate. As shown in Fig.~\ref{fig:spe_idv3}, residuals remain near the detection threshold throughout most of the time window, with several models exhibiting small oscillatory spikes near the end, which gradually diminish in amplitude.

Back-to-control faults (Faults 4, 5, and 7) are reliably detected by all models except in the case of Fault 5, where performance varies sharply. EfficientKAN-AE achieves perfect detection, followed closely by FastKAN-AE, while OAE and FourierKAN-AE detect fewer than 40\% of fault occurrences. As shown in Fig. \ref{fig:spe_idv5}, only the EfficientKAN-AE and FastKAN-AE models maintain SPE values above the detection threshold after the control system restores the process variables to nominal conditions. This suggests that the adaptive basis functions in KAN-AEs help them capture subtle deviations affecting only one or a few process variables, even when trained with limited data.

Uncontrollable faults exhibit greater variation in model performance. While all methods detect clearly separable faults such as 6 and 14, performance deteriorates on faults like 10, 16, and 19, particularly when the fault signature is intermittent. In these cases, SPE values often dip below the detection threshold during the fault window, leading to missed detections (see Fig. \ref{fig:spe_idv16}). This pattern reflects the inability of static autoencoders to capture temporal dependencies or respond consistently to delayed or fluctuating fault signatures. The limitation is not specific to KAN-AEs and applies broadly to models lacking explicit temporal structure. In such cases, recurrent architectures such as RNNs or LSTMs may offer improved robustness. Despite this, EfficientKAN-AE outperforms other variants on each of these faults, surpassing the next best models by 13–34 percentage points.

\begin{table}[t]
\centering
\caption{Fault Detection Rate (FDR) per fault for each model (500 training samples). Values are reported as mean $\pm$ 95\% confidence interval, expressed as percentages.}
\label{tab:fdr_500}
\scriptsize
\resizebox{\linewidth}{!}{%
\begin{tabular}{c|c|c|c|c|c}
\toprule
\textbf{Fault} & \textbf{OAE} & \textbf{EfficientKAN-AE} & \textbf{FastKAN-AE} & \textbf{FourierKAN-AE} & \textbf{WavKAN-AE} \\
\midrule
 3  & $6.25 \pm 0.61$ & $5.98 \pm 0.55$ & $5.64 \pm 0.85$ & $5.57 \pm 0.80$ & $\mathbf{4.69 \pm 0.83}$ \\
 9  & $5.90 \pm 0.58$ & $6.96 \pm 0.89$ & $5.24 \pm 0.82$ & $4.74 \pm 0.79$ & $\mathbf{3.76 \pm 0.90}$ \\
 15 & $8.72 \pm 0.89$ & $8.99 \pm 0.83$ & $\mathbf{7.24 \pm 0.76}$ & $9.25 \pm 0.69$ & $9.22 \pm 0.81$ \\
\midrule
 4  & $\mathbf{99.91 \pm 0.10}$ & $99.56 \pm 0.20$ & $95.58 \pm 2.91$ & $75.68 \pm 9.64$ & $99.78 \pm 0.29$ \\
 5  & $36.62 \pm 1.32$ & $\mathbf{100.00 \pm 0.00}$ & $96.51 \pm 3.82$ & $30.41 \pm 1.20$ & $87.87 \pm 6.58$ \\
 7  & $99.99 \pm 0.02$ & $\mathbf{100.00 \pm 0.00}$ & $\mathbf{100.00 \pm 0.00}$ & $\mathbf{100.00 \pm 0.00}$ & $\mathbf{100.00 \pm 0.00}$ \\
\midrule
 1  & $\mathbf{99.86 \pm 0.03}$ & $99.66 \pm 0.03$ & $99.53 \pm 0.03$ & $99.57 \pm 0.05$ & $99.65 \pm 0.05$ \\
 2  & $98.59 \pm 0.10$ & $\mathbf{98.60 \pm 0.07}$ & $98.11 \pm 0.09$ & $98.50 \pm 0.06$ & $98.47 \pm 0.03$ \\
 6  & $\mathbf{100.00 \pm 0.00}$ & $\mathbf{100.00 \pm 0.00}$ & $\mathbf{100.00 \pm 0.00}$ & $\mathbf{100.00 \pm 0.00}$ & $\mathbf{100.00 \pm 0.00}$ \\
 8  & $98.10 \pm 0.11$ & $98.07 \pm 0.14$ & $\mathbf{98.19 \pm 0.09}$ & $98.00 \pm 0.13$ & $98.00 \pm 0.17$ \\
 10 & $59.89 \pm 2.37$ & $\mathbf{83.66 \pm 1.24}$ & $66.57 \pm 1.49$ & $47.48 \pm 1.20$ & $70.15 \pm 2.10$ \\
 11 & $\mathbf{78.18 \pm 1.16}$ & $74.83 \pm 1.46$ & $72.98 \pm 1.65$ & $67.03 \pm 3.63$ & $77.56 \pm 2.18$ \\
 12 & $98.86 \pm 0.08$ & $\mathbf{99.34 \pm 0.06}$ & $99.03 \pm 0.04$ & $98.74 \pm 0.12$ & $99.00 \pm 0.09$ \\
 13 & $\mathbf{95.38 \pm 0.08}$ & $95.12 \pm 0.13$ & $94.83 \pm 0.08$ & $94.58 \pm 0.11$ & $94.50 \pm 0.07$ \\
 14 & $99.99 \pm 0.01$ & $\mathbf{100.00 \pm 0.00}$ & $99.98 \pm 0.02$ & $99.98 \pm 0.02$ & $\mathbf{100.00 \pm 0.00}$ \\
 16 & $55.26 \pm 2.55$ & $\mathbf{85.57 \pm 1.66}$ & $63.28 \pm 1.66$ & $36.00 \pm 1.35$ & $67.53 \pm 1.94$ \\
 17 & $96.15 \pm 0.29$ & $\mathbf{96.67 \pm 0.28}$ & $95.22 \pm 0.36$ & $89.66 \pm 1.34$ & $93.55 \pm 0.79$ \\
 18 & $\mathbf{90.57 \pm 0.17}$ & $90.24 \pm 0.17$ & $90.00 \pm 0.21$ & $90.11 \pm 0.27$ & $90.13 \pm 0.22$ \\
 19 & $33.30 \pm 2.43$ & $\mathbf{77.40 \pm 3.36}$ & $32.07 \pm 4.52$ & $14.71 \pm 2.57$ & $43.79 \pm 8.25$ \\
 20 & $62.27 \pm 1.19$ & $\mathbf{71.68 \pm 1.07}$ & $58.32 \pm 1.51$ & $50.85 \pm 1.76$ & $59.37 \pm 2.40$ \\
 21 & $46.81 \pm 1.82$ & $\mathbf{49.17 \pm 2.53}$ & $47.46 \pm 1.49$ & $43.85 \pm 1.35$ & $44.23 \pm 1.39$ \\
\bottomrule
\end{tabular}
}
\end{table}

\begin{table}[h]
\centering
\caption{False Alarm Rate (FAR) per fault for each model (500 training samples). Values are reported as mean $\pm$ 95\% confidence interval, expressed as percentages.}
\label{tab:far_500}
\scriptsize
\resizebox{\linewidth}{!}{%
\begin{tabular}{c|c|c|c|c|c}
\toprule
\textbf{Fault} & \textbf{OAE} & \textbf{EfficientKAN-AE} & \textbf{FastKAN-AE} & \textbf{FourierKAN-AE} & \textbf{WavKAN-AE} \\
\midrule
 3  & $3.08 \pm 0.95$ & $\mathbf{6.02 \pm 1.52}$ & $3.31 \pm 1.82$ & $2.67 \pm 1.48$ & $3.46 \pm 1.74$ \\
 9  & $\mathbf{9.21 \pm 1.13}$ & $\mathbf{8.54 \pm 1.33}$ & $\mathbf{6.13 \pm 1.36}$ & $\mathbf{8.19 \pm 1.28}$ & $\mathbf{8.17 \pm 1.62}$ \\
 15 & $2.04 \pm 0.64$ & $2.77 \pm 0.60$ & $1.46 \pm 0.53$ & $1.10 \pm 0.35$ & $0.44 \pm 0.20$ \\
\midrule
 4  & $2.00 \pm 0.40$ & $1.60 \pm 0.57$ & $1.56 \pm 0.44$ & $1.62 \pm 0.48$ & $1.60 \pm 0.45$ \\
 5  & $2.00 \pm 0.40$ & $1.60 \pm 0.57$ & $1.56 \pm 0.44$ & $1.62 \pm 0.48$ & $1.60 \pm 0.45$ \\
 7  & $1.90 \pm 0.52$ & $2.00 \pm 0.68$ & $1.15 \pm 0.55$ & $1.27 \pm 0.39$ & $1.04 \pm 0.40$ \\
\midrule
 1  & $2.19 \pm 0.40$ & $2.48 \pm 0.55$ & $1.12 \pm 0.63$ & $0.92 \pm 0.37$ & $1.06 \pm 0.35$ \\
 2  & $1.13 \pm 0.35$ & $1.31 \pm 0.54$ & $1.69 \pm 0.58$ & $1.21 \pm 0.38$ & $1.65 \pm 0.48$ \\
 6  & $1.00 \pm 0.29$ & $1.29 \pm 0.42$ & $1.54 \pm 0.67$ & $1.31 \pm 0.31$ & $1.15 \pm 0.34$ \\
 8  & $2.96 \pm 0.55$ & $3.75 \pm 0.91$ & $4.29 \pm 1.48$ & $3.35 \pm 1.62$ & $2.27 \pm 1.50$ \\
 10 & $1.77 \pm 0.44$ & $1.04 \pm 0.30$ & $1.46 \pm 0.42$ & $1.00 \pm 0.31$ & $0.81 \pm 0.34$ \\
 11 & $2.67 \pm 0.55$ & $2.04 \pm 0.52$ & $2.71 \pm 0.69$ & $1.52 \pm 0.51$ & $0.63 \pm 0.37$ \\
 12 & $3.35 \pm 0.77$ & $3.75 \pm 0.99$ & $4.12 \pm 1.34$ & $3.35 \pm 1.13$ & $2.90 \pm 1.38$ \\
 13 & $1.29 \pm 0.40$ & $1.04 \pm 0.39$ & $1.35 \pm 0.63$ & $0.85 \pm 0.33$ & $0.65 \pm 0.26$ \\
 14 & $2.06 \pm 0.54$ & $1.00 \pm 0.32$ & $1.54 \pm 0.52$ & $0.88 \pm 0.31$ & $0.48 \pm 0.25$ \\
 16 & $\mathbf{9.65 \pm 1.43}$ & $\mathbf{7.60 \pm 1.38}$ & $\mathbf{20.02 \pm 2.20}$ & $\mathbf{14.75 \pm 2.37}$ & $\mathbf{12.90 \pm 2.83}$ \\
 17 & $3.81 \pm 0.89$ & $3.73 \pm 1.06$ & $1.90 \pm 0.64$ & $2.02 \pm 0.56$ & $1.71 \pm 0.53$ \\
 18 & $2.50 \pm 0.68$ & $2.31 \pm 0.58$ & $1.73 \pm 0.65$ & $1.33 \pm 0.49$ & $0.65 \pm 0.30$ \\
 19 & $1.35 \pm 0.45$ & $2.52 \pm 0.63$ & $2.73 \pm 0.85$ & $0.75 \pm 0.32$ & $0.42 \pm 0.31$ \\
 20 & $0.96 \pm 0.39$ & $0.65 \pm 0.25$ & $1.00 \pm 0.50$ & $0.44 \pm 0.23$ & $0.48 \pm 0.24$ \\
 21 & $\mathbf{5.17 \pm 1.21}$ & $\mathbf{5.35 \pm 1.22}$ & $2.92 \pm 1.07$ & $2.92 \pm 1.08$ & $2.42 \pm 1.06$ \\
\bottomrule
\end{tabular}
}
\end{table}

\subsubsection{False Alarm Rates}
\begin{figure*}[t]
    \centering
    \begin{subfigure}[t]{0.31\textwidth}
        \centering
        \includegraphics[width=\textwidth]{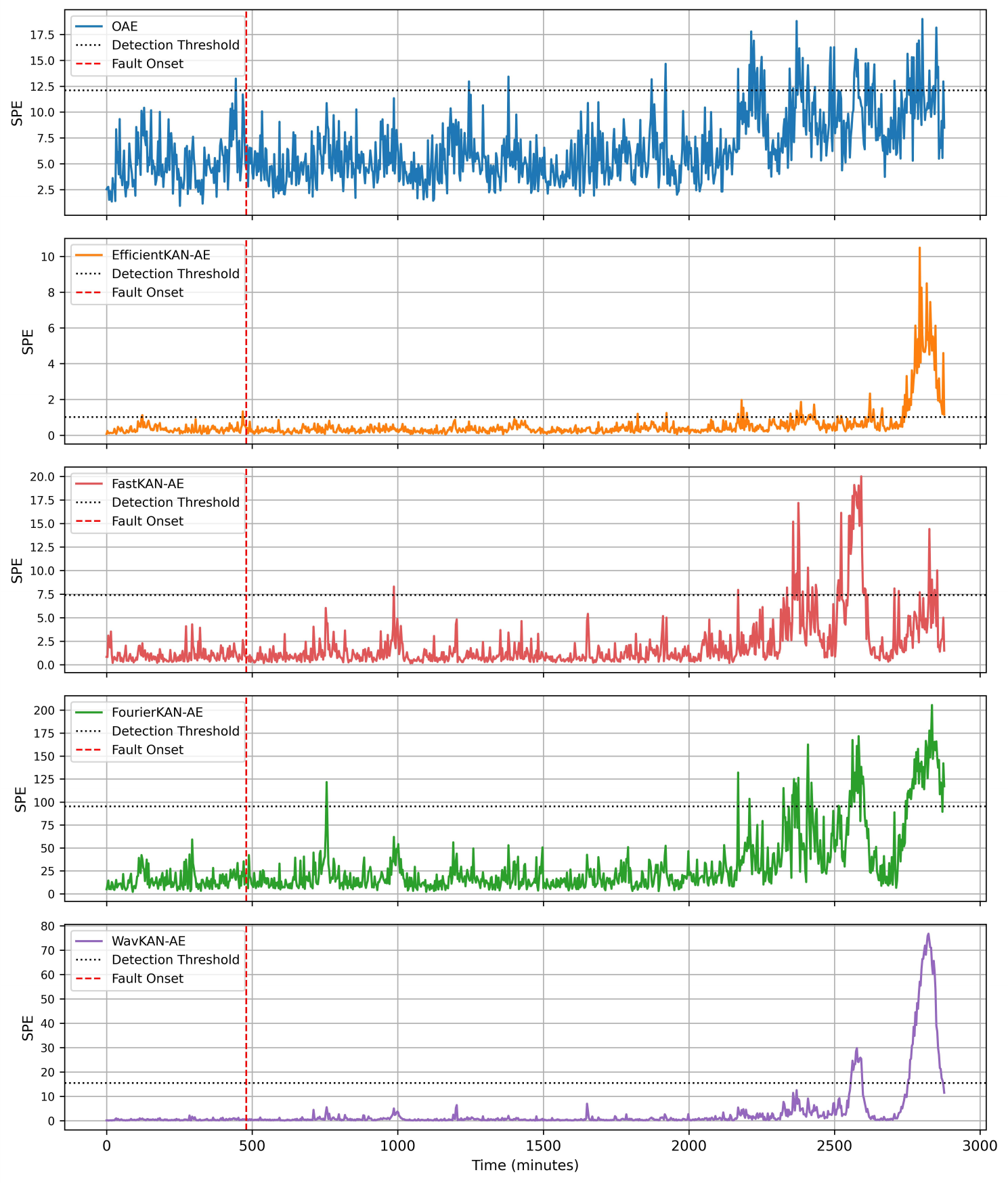}
        \caption{Fault 3}
        \label{fig:spe_idv3}
    \end{subfigure}
    \hfill
    \begin{subfigure}[t]{0.31\textwidth}
        \centering
        \includegraphics[width=\textwidth]{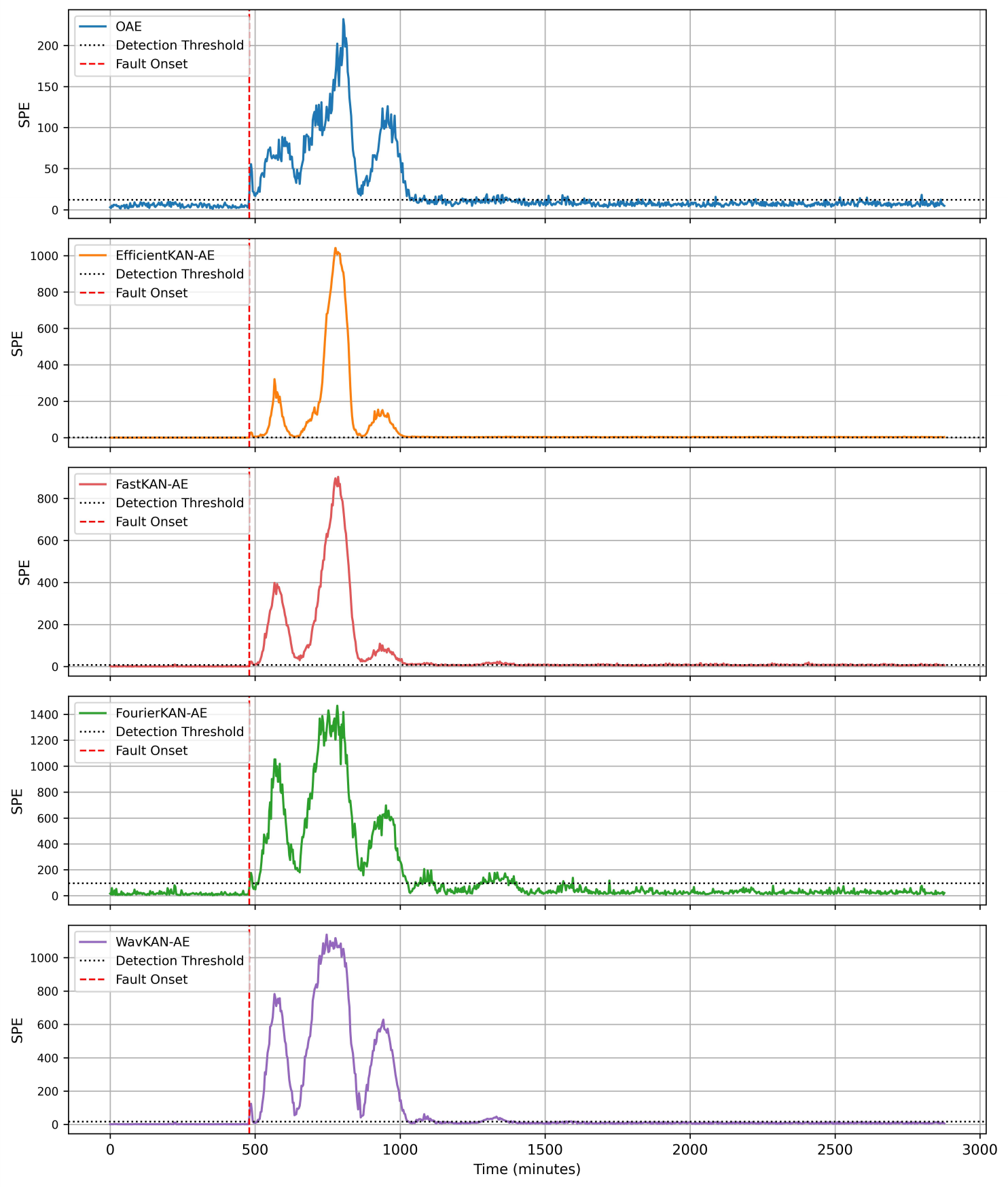}
        \caption{Fault 5}
        \label{fig:spe_idv5}
    \end{subfigure}
    \hfill
    \begin{subfigure}[t]{0.31\textwidth}
        \centering
        \includegraphics[width=\textwidth]{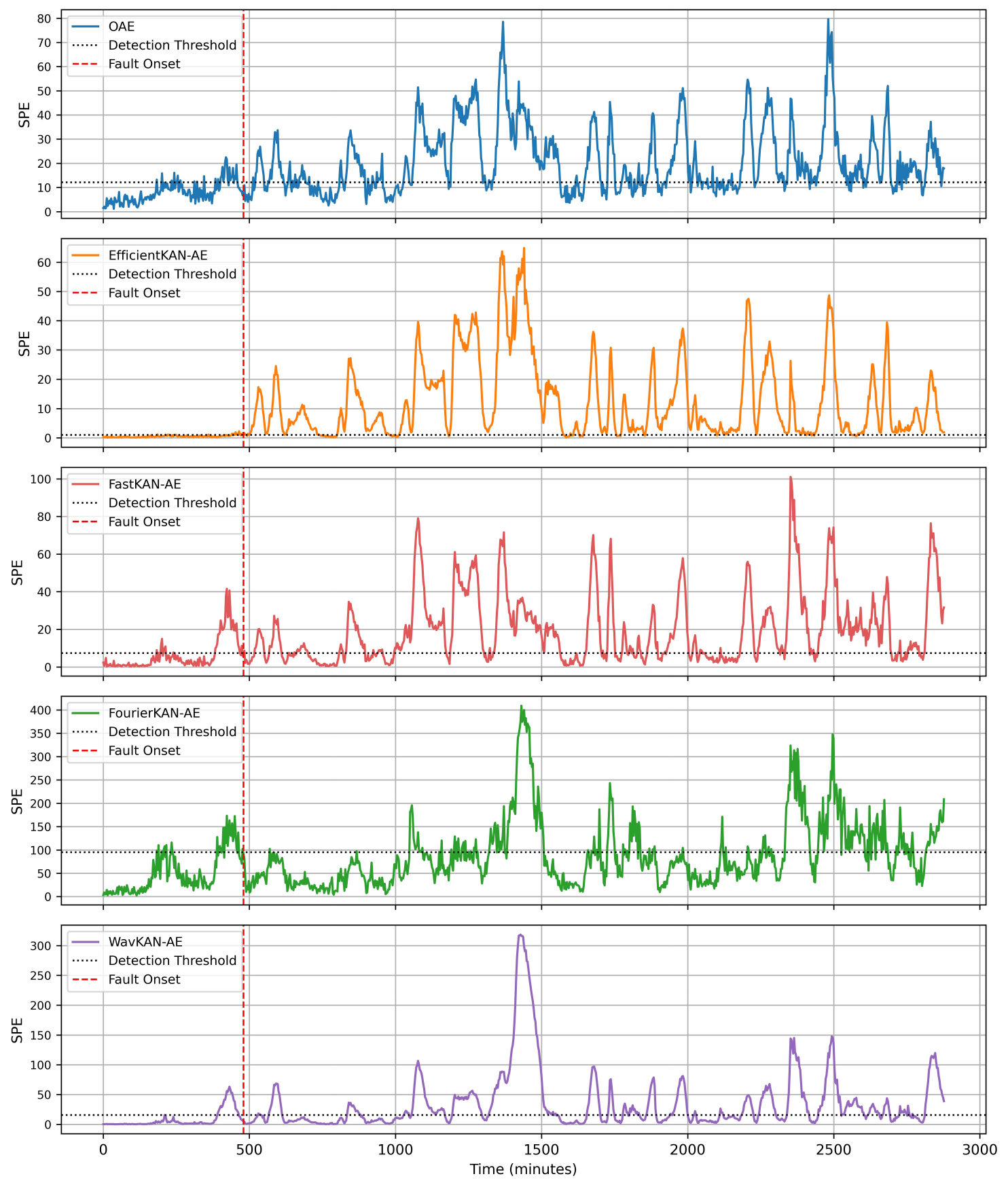}
        \caption{Fault 16}
        \label{fig:spe_idv16}
    \end{subfigure}
    
    \caption{Temporal evolution of SPE values across AE variants under Faults 3, 5, and 16. All models were trained on 500 fault-free samples.}
    \label{fig:spe_idv_selected}
\end{figure*}

To assess whether high detection rates reflect true fault sensitivity rather than excessive triggering, we examine the false alarm rates (FARs) in Table~\ref{tab:far_500}. Most models remain below the 5\% threshold for the majority of faults, indicating that detection is not driven by over-sensitivity, despite training with limited data. However, Fault 16 stands out across models, with all variants exhibiting pre-fault rises in reconstruction error (Fig.~\ref{fig:spe_idv16}), likely contributing to the elevated false alarm rates. Among them, EfficientKAN-AE contains this response more tightly, while others produce prolonged or irregular spikes.

\section{Conclusions}
\label{sec:conclusions}
This work conducted a comparative evaluation of Kolmogorov–Arnold autoencoders (KAN-AEs) for unsupervised fault detection using an MLP-based orthogonal autoencoder as the baseline, examining how different edge function parameterizations (e.g., B-splines, wavelets) affect detection performance across different training set sizes. Performance differences were validated through the Bayesian signed-rank test, offering more robust statistical inference than traditional frequentist approaches. Results show that KAN-AEs can match or surpass the MLP-based baseline (OAE) while requiring substantially less training data. WavKAN-AE achieved top performance with only 13\% of the data needed by the OAE (4,000 compared to 30,500 samples), demonstrating both scalability and data efficiency. EfficientKAN-AE reached peak performance with just 3\% of the OAE's training requirement (1,000 compared to 30,500 samples), dominating in the data-scarce regime. FastKAN-AE required approximately 69\% more training data than the OAE (51,500 compared to 30,500 samples) to reach comparable performance. In contrast, FourierKAN-AE consistently underperformed, highlighting the limitations of global basis functions in capturing the underlying data features that enable fault detection.

These findings suggest that the choice of edge function plays a central role in model generalization and data efficiency. The architectural simplicity and parameter efficiency of KAN-AEs, combined with their robustness in data-scarce regimes, make them strong candidates for deployment in industrial monitoring scenarios where fault data are limited or costly to obtain.

However, the models evaluated in this study do not capture temporal dependencies and are restricted to fault detection, without addressing fault identification or root-cause diagnosis. The lack of temporal modeling particularly limits their ability to detect intermittent or evolving fault patterns. Additionally, while KANs are often presented as interpretable due to their functional edge representations, this aspect is not systematically evaluated in our current framework. A more rigorous analysis is needed to assess whether KANs’ structural transparency translates into actionable insights for process monitoring.

To address these limitations, future research could explore several directions. One promising avenue is the integration of temporal modeling within KAN-AEs using recurrent architectures, temporal convolutions, or latent state-space formulations to enable sequential fault detection and better capture delayed or evolving fault patterns. Additionally, future work should develop interpretability tools that help operators understand how KAN-based models make decisions, going beyond standard contribution plots to provide actionable diagnostic insights.

\section{Acknowledgments}
This work was supported by the McMaster Advanced Control Consortium (MACC).
\section{CRediT authorship contribution statement}
Enrique Luna Villagómez: Conceptualization, Methodology, Software, Writing – original
draft. Vladimir Mahalec: Supervision, Conceptualization, Methodology, Writing – review \&
editing.

\bibliographystyle{elsarticle-num} 
\bibliography{cas-refs}






\end{document}